\documentclass{article}
\usepackage[utf8]{inputenc}
\usepackage[notes,backend=biber]{biblatex-chicago}
\usepackage[american]{babel}
\usepackage{csquotes}
\usepackage{hyperref}
\usepackage{csquotes}
\usepackage{url}
\hypersetup{
    colorlinks=true,
    linkcolor=black,
    filecolor=magenta,      
    urlcolor=blue,
    citecolor=black,
    pdfpagemode=FullScreen,
}
\usepackage{enumerate,amsmath,amssymb}
\usepackage{setspace}

\title{%
  Contextual Trust \\
  \large Essay Submission for Undergraduate Honors \\
  \small Stanford University Department of Philosophy}
\author{Ryan Othniel Kearns}
\date{May 9, 2022}

\begin{document}

\maketitle
\tableofcontents
\newpage

\section*{Acknowledgements}
I am indebted to Thomas Icard, Kathleen Creel, Nadeem J.Z. Hussain, Karina Sahgal, Christopher Bobonich, Hyoung Sung Kim, Jared Warren, Johan van Benthem, John Wilcox, Samantha Els, Josh Nkoy, Benjamin Wittenbrink, Ben Sparkes, Matthew Hall, Philip Pfeffer, Cameron McClellan, and others unnamed for their guidance and support of this project.

\newpage

\section{Abstract}

Trust is an important aspect of human life. It provides instrumental value in allowing us to collaborate on and defer actions to others, and intrinsic value in our intimate relationships with romantic partners, family, and friends.\autocite{sep-trust} In this paper I examine the nature of trust from a philosophical perspective. Specifically I propose to view trust as a context-sensitive state in a manner that will be made precise. I am interested in instances of trust where certain contextual conditions are relevant. The contribution of this paper is threefold.

First, I make the simple observation that an individual's trust is typically both action- and context-sensitive. Action-sensitivity means that trust may obtain between a given truster and trustee for only certain actions. Context-sensitivity means that trust may obtain between a given truster and trustee, regarding the same action, in some conditions and not others. This observation is not commonly confronted in existing philosophical theories about trust,\footnote{See \cite{baier_1986}, \cite{jones-affective-attitude}, or \cite{Nguyen-trust} for influential accounts that omit context-sensitivity. I give the analysis to back this claim in the literature review, especially section \ref{TrustInPhilosophy}.} leading to confusion about the nature of trust. Explicitly including context in our formulation of trust can clarify our understanding. We replace the general question ``What does it mean to trust?'' with a more specific and tractable one, ``What does it mean for $A$ to trust $B$ to do $X$ in context $C$?'' We will be precise about what constitutes a context, and what kinds of facts get to count as contextual factors. I also opine about what kinds of things may play the role of the truster $A$, trustee $B$, and action $X$.

Second, I advance a theory for the nature of contextual trust. I propose that the answer to ``What does it mean for $A$ to trust $B$ to do $X$ in context $C$?'' has two conditions. First, $A$ must take $B$'s doing $X$ as a means towards one of $A$'s ends. Second, $A$ must adopt an unquestioning attitude concerning $B$'s doing $X$ in context $C$. This unquestioning attitude is similar to the attitude introduced in Nguyen 2021, but with subtle alterations.

Finally, we explore how contextual trust can help us make sense of trust in general non-interpersonal settings, notably that of artificial intelligence (AI) systems. The field of Explainable Artificial Intelligence (XAI) assigns paramount importance to the problem of user trust in opaque computational models, yet does little to give trust diagnostic or even conceptual criteria. I propose that contextual trust is a natural fit for the task by illustrating that model transparency and explainability map nicely into our construction of the contexts $C$.

\newpage
\section{Introduction}\label{Introduction}

\subsection{Trusting in Context}\label{TrustingInContext}
My English friend recently learned to drive in the United States. Once he received his license, I began lending him my car so that he could practice. I did not always have time to go with him, and given his new license I trusted him to handle the car on his own. This experiment ended in near-disaster after he drove the wrong way down a one-way street in San Francisco, nearly crashing into oncoming traffic in the process. As it turns out, my friend's license was obtained in a suburban neighborhood near our university, and had not prepared him for the intricacies of city driving. Now, our policy has changed. My friend can still drive my car around our neighborhood unsupervised. I do not trust him to drive in the city, though. So, when he wants to practice this particular skill, I make sure to be in attendance to point out and correct any major errors.

This state of trust I have towards my friend is complex. It is not enough to say I trust him to drive my car sometimes, but not other times. The state I am in is more precise. There are boundary conditions, contextual conditions, where my trust runs out. If queried, I could define them specifically -- my friend can drive my car alone in suburbia, but not in the city, not during a torrential rainstorm, and not while blindfolded. Also, I expect the conditions for my trust will relax in time, proportional to his demonstrated proficiency at one-way streets, unprotected left turns, \textit{et cetera}. Soon I may trust my friend to drive in the city and in the rain, though still not while blindfolded. In fact, my trust in my friend has already changed in this way -- it did so both when he received his license, and then again after the one-way street incident.

Mainstream philosophical accounts of trust that take trust to be a tripartite relation involving trusters, trustees, and actions cannot define this trust in a satisfiable way. The tripartite question, ``Do you trust your friend to drive your car?,'' is not specific enough. My answer depends on contextual factors, including my friend's license, the location where the driving will occur, and my friend's track record of driving in said location. These factors are \textit{difference-making} in my decision to trust, and we cannot account for them in speaking just of trusters, trustees, and actions.\footnote{Some may object that a suitably fine-grained view of actions could capture all difference-making contextual factors. I present this objection in section \ref{ActionGranularity} and lay down some reasons to resist it. For now, it should be clear that phrases like ``drive a car'' are at least valid action phrases, and if we want to discuss them we ought to introduce difference-making contextual factors.}

My point in this example is to demonstrate that we can talk of trust more richly than contemporary philosophical theory seems to allow. We will use \textbf{contextual trust} to denote this view of trust as a four-place relation, including some kind of context as the fourth argument. In speaking of contextual trust we gain much in expressive power, and can elucidate some important cases of trust in modern usage. So I will argue. But first, why are we talking about trust in the first place?

\subsection{The Ubiquity of Trust}\label{UbiquityOfTrust}
Trust is all around us. If you ordered coffee this morning, you have already trusted a barista not to poison your drink. Of course, this example is outlandish, and most actual cases of trust are mundane. We trust fellow drivers to use their indicators responsibly, waiters to properly record our orders, and friends to keep embarrassing secrets. Annette Baier writes that we ``inhabit a climate of trust as we inhabit an atmosphere.''\autocite[234]{baier_1986} Like an atmosphere, trust can diffuse into our immediate surroundings. When we walk alone at night, or leave our grocery cart unattended, Margaret Urban Walker observes, it seems we trust no one in particular except simply ``people.''\autocite{walker_2006}

Not only can trust be unavoidable in this way, but we often seem to enter into trusting others unwittingly. Baier continues to say we ``notice it as we notice air, only when it becomes scarce or polluted.''\autocite[234]{baier_1986} Perhaps the comparison doesn't always work. I certainly notice when trusting my belayer to catch my fall. The point is taken, though. In the majority of everyday cases we seem to not notice ourselves trusting. Most all of us grew up trusting our immediate caretakers for our very survival. It seems unlikely such trust is ever consciously evaluated. We might trust our doctors, our news anchors, and our professors without ever noticing we are doing so. Some people trust law enforcement agents without a second thought; others cannot. As Walker observes, we ought to feel fortunate when our lives aren't riddled with skepticism about cases like these.

It also appears that we utilize trust for a variety of reasons. Trust lets us divide labor, such as when we trust project partners to tackle sections of a presentation. Trust in others lets us perform unilateral actions we could not wisely undertake otherwise, like protected turns at intersections. And, in simple yet important fashion, trust is essential for certain lasting positive relationships with others, such as parents, children, and romantic partners.

This ubiquity and diversity of trust in daily life makes the philosophical analysis of the concept a thorny task. Baier notes that we trust ``wisely or stupidly'' in ``a great variety of forms... both with intimates and with strangers.''\autocite[234]{baier_1986} Walker, discussing her own conception of trust in the book \textit{Moral Repair}, says:
\begin{displayquote}
    ``...it is tempting to load a great deal into an account of trust, and so to moralize it, personalize it, or narrow it. Doing so can produce rich accounts of particular kinds and circumstances of trust, but will fail to comprehend in a single account all the varieties of trust there are.''\autocite[75]{walker_2006}
\end{displayquote}
Walker's point highlights an irreconcilable compromise for anyone exploring the nature of trust philosophically. Rich and descriptive accounts of the concept will invariably leave some cases out. Generic and inclusive accounts can do better at picking up edge cases. Yet, it seems likely such approaches will either count some definite cases of non-trust as trust, or else provide a vapid definition with little explanatory power.

Nonetheless, there is a rich philosophical literature on this topic. Before diving into the literature, though, the following sections \ref{TrustAndVulnerability} and \ref{WhyTrustAtAll} will shed light on some common through lines in existing theories. Most philosophical theories of trust tend to accommodate and explain these following features.


\subsection{Trust and Vulnerability}\label{TrustAndVulnerability}

Trust has many attractive features, making it a desirable state. Trust also has an intimate connection to vulnerability of a particular kind, though. When I trust you, I put myself at risk relative to you. We talk of one \textit{abusing} or, tellingly, \textit{taking advantage of} another's trust. Our language suggests that trusting can be a dangerous state. When we trust, we anticipate certain actions by other people, and base our own agency off of those actions. We take on an ``accepted vulnerability.''\autocite[235]{baier_1986} Thus, we expose ourselves to the disappointment and risk that comes when those actions are not completed as expected.\autocite[78]{walker_2006}

The notion of ``Zero Trust'' in cybersecurity is illustrative of this relationship.\footnote{\cite{nist_zero_trust}} A Zero Trust Architecture is a network security architecture granting minimal permissions to devices until they pass strong authentication checks. Zero Trust operates with the adage ``never trust, always verify,'' and defines the goals of ``eliminating implicit trust'' and ``continuously validating'' network behavior.\autocite{zero_trust_architecture} The phrase ``never trust, always verify'' implies that verification can be substituted for trust, at least in a cybersecurity context.

An adjacent field, cryptocurrency, also takes heavy computational pains to eliminate any need for trust. In the seminal Bitcoin whitepaper from 2008, Satoshi Nakamoto concludes with the phrase, ``We have proposed a system for electronic transactions without relying on trust.''\autocite{nakamoto_2008} Cryptocurrencies run atop blockchains, immutable and publicly viewable digital ledgers. Blockchains validate user transactions in a peer-to-peer network with a computational system called ``proof-of-work,'' which prohibits rogue actors from making fraudulent transactions so long as they hold a minority in the network. Nakamoto's design replaces trust in a central bank with trust in mathematical proof and computer networking.

The trust-verification tradeoff is also applicable beyond computing. The non-trusting parent may check the security cameras on a weekend away. James Bond checks his guest suite for bugs. Such examples specify that trusting involves, at base, a kind of \textit{epistemic vulnerability}. We may trust where we lack certainty, and when we fail to trust, we seek some kind of verification. While other vulnerabilities surely apply -- trusting a cyclist to stop in front of us incurs physical danger -- they arise out of a more basic epistemic shortcoming. If I was \textit{certain} the cyclist would stop, I would not report to be in danger.

\subsection{Why Trust at All?}\label{WhyTrustAtAll}
Cybersecurity professionals, cryptocurrency advocates, and the like may raise the question -- why trust at all? In settings where total verification is possible, it indeed may seem that trust is unnecessary. Yet I will argue there are at least two reasons why trust must generally be counted on. The first argument concerns the \textit{necessity} for trust in everyday life, and the second concerns trust's \textit{value} beyond the mere epistemic utility mentioned above.

\subsubsection{The Necessity of Trust}
The first reason for needing trust has to do with \textbf{information overload}. Going about our daily lives, we cannot hope to verify even a fraction of our assumptions. Many such assumptions are indispensable. I may never fully understand how my carburetor functions, how my dentist reads X-rays, or how my sausages are made. Yet I drive to work, wear braces, and eat breakfast all the same. Doing so only after verification would be preposterous; such a policy would consume all of my time. Thus, we are forced into trust by our finite time and limited mental bandwidth. We must act without complete evidence, and in doing so, we trust. C. Thi Nguyen says colorfully that trust is ``part of a reasonable strategy for coping with the cognitive onslaught of the world.''\autocite[2]{Nguyen-trust}

\subsubsection{The Value of Trust}\label{ValueOfTrust}
Trust is also needed since epistemic certainty does not succeed in explaining trust's \textbf{value}. There are two ways in which this can occur.

The first way has to do with a phenomenon called \textbf{therapeutic trust}. Often times, we trust without expectation that our trust will be fulfilled. In these cases our reasons for trusting extend beyond the action being trusted. Victoria McGeer talks of parents lending the house or family car to a teenager, expecting that their trust in the teenager could be violated.\autocite[241]{McGeer-trust-empowerment} In this case, the parents' trust does not stand in for epistemic certainty, but instead has the long-term goal of eliciting responsible behavior. In a similar vein, Richard Holton talks of shop owners hiring ex-petty criminals and trusting them, not because of any demonstrated reliability, but in order to draw them back into the moral community.\autocite[63]{Holton1994-HOLDTT}

The second way has to do with a longstanding distinction in the philosophy of trust, dating back to 1986 with Baier,\autocite{baier_1986} and having to do with the difference between trust and reliance. Baier distinguishes trust from ``mere reliance'' by the possibility of betrayal. When our trust is violated, we are liable to feel betrayed. Yet, when we were merely relying on someone, we can be disappointed by their failure to act, but not justifiably betrayed. Richard Holton has said similarly that trust is reliance plus investment in the right kinds of \textit{reactive attitudes}\autocite[67]{Holton1994-HOLDTT} -- specifically, the readiness to feel betrayal when trust is let down. Baier and Karen Jones interpret this divide between trust and reliance as indicating normative expectations of the trustee by the truster. In relying, you might treat someone as a ``mere regularity to be worked with or around,''\autocite[669]{Jones2015} but when you trust, you treat the same person as a fellow agent, subject to normative and not just predictive expectations.

So trust is a ubiquitous property in our social lives, we require it in everyday life to overcome information overload, and our reasons for trusting seem more complex than merely seeking reliability. Clearly we cannot do away with trust in philosophy as we do in cybersecurity. ``Never trust, always verify'' would be a disastrous policy in our interactions with spouses, professors, car mechanics, and cashiers. In such cases, trust is a more complex attitude or relation than reliance or certainty. In what follows in this paper, I will discuss how some philosophers have attempted to proceed beyond these baseline intuitions.

\subsection{Approach}
Recall that section \ref{UbiquityOfTrust} included the important observation from Margaret Urban Walker that philosophical theories of trust must balance precision with generality. Given the variety of settings where trust seems to be relevant, it is difficult to have both qualities in a single theory.

My goal with this paper, in giving an account of the nature of trust, is to strike a balance between these two extremes. In section \ref{DifferenceMakingFeatures}, I replace the original, vague, and slippery question -- ``What does it mean to trust?'' -- with something more specific. I will argue that a natural substitute is a question like, ``What does it mean for $A$ to trust $B$ to do action $X$ in context $C$?'' The types of things that can stand in for $A$, $B$, $X$, and $C$ will be made precise in sections \ref{Directedness}, \ref{ActionSensitivity}, and \ref{ContextSensitivity}. Note that this approach differs from the mainstream philosophy of trust, which typically takes only $A$, $B$, and $X$ into account. In section \ref{ContextSensitivity} I show how including $C$ can make our understanding of trust more precise. In section $\ref{ContextualTrustInPreviousTheories}$ I discuss how other contemporary philosophical theories already incorporate contextual factors, just under different names.

After getting more specific about our question, I will provide my own theoretical answer in section \ref{DefiningContextualTrust}. My answer draws inspiration from philosophers Jeff Buechner and Herman Tavani,\autocite{buechner-tavani} Richard Holton,\autocite{Holton1994-HOLDTT} Karen Jones, \autocite{jones-affective-attitude} C. Thi Nguyen,\autocite{Nguyen-trust} and Margaret Urban Walker,\autocite{walker_2006} but does not conform precisely to any of their past work.

These results may be decoupled. One may believe that trust is contextual like I describe, yet goodwill-oriented like Baier, Jones, or Cogley describe.\autocite{baier_1986,jones-affective-attitude,Cogley2012-COGTAT} In other words, readers convinced of my points about context need not believe in my answer to ``What does it mean for $A$ to trust $B$ to do action $X$ in context $C$?'' Yet I intend to show my answer has some attractive features.

In section \ref{RepliesToObjections}, I defend my generic theory from objections, and motivate the case that it accurately characterizes trust. In section \ref{CaseStudyXAI}, I apply this theory to the case study of Explainable Artificial Intelligence (XAI). XAI seeks to increase trust in AI models with precise explanations of model behavior, yet never precisely defines trust itself. A theory of trust incorporating context like this can thus explain how XAI concepts like model transparency and explainability influence trust. Such a result indicates that a context-oriented theory of trust can have useful applications both in and outside of philosophy.

\section{Trust in prior literature}\label{Background}
As may be expected by now, trust has a rich literary history in fields beyond just philosophy. In what follows, we first examine trust in some relevant, adjacent disciplines, and then dive specifically into philosophical theory.

\subsection{Trust in Economics}\label{Economics}
Economists tend to view trust under a family of ``expected outcome'' models, or what philosophers might call rationalist accounts of trust. Bhattacharya et al. (1998) propose a formal theory of trust that bridged then-existing theories from psychology and economics.\autocite{bhattacharya} According to the authors, many existing psychological theories\autocite{lewicki-bunker,kreps_1990} emphasize inherent trustworthiness as a personality trait, which downplays the situation-specificity of trust for economic settings. At the same time, economic theories\autocite{Dasgupta1988-DASTAA} are overly concerned with situation-specific constraints and do not consider interpersonal relationships in transactions where trust is important. So, a composite mathematical theory must encode both aspects of the situation and the trust-giving and -receiving parties.

The authors of this paper purport to do this. For a sequential scenario, in which agent 1 acts before agent 2, agent 1's trust of agent 2, $T_{1,2}$, can be given as:
\begin{align*}
    T_{1,2}|a_1^* &= Pr(\mu_1 > 0 | a_1^*) \\ &= \sum_{x\in\gamma_1}Pr(\alpha_1 = x | a_1^*) \\
    &= \sum_{x\in\gamma_1}\sum_{a_2\in A_2} F_1(x_1; a_1^*, a_2)\cdot c_1(a_2|a_1^*)
\end{align*}
where:
\begin{itemize}
    \item $a_1^*$ is the action agent 1 decides to take.
    \item $\mu_1$ is the ``goodness" of an outcome $x_1$, and when $\mu_1 > 0$ the outcome is favorable for agent 1.
    \item $\alpha_1$ is the function taking an action ($a_1$) to an outcome ($x_1$). In a special case it is determinate, but we assume randomness in general cases.
    \item $\gamma_1$ is the set of all potential outcomes of some action $a_1$, the codomain of $\alpha_1$.
    \item $A_2$ is the set of all actions that agent 2 may take after agent 1 has taken their decided action ($a_1^*$).
    $F_1$ is the function taking joint actions from agents 1 and 2 to outcomes for agent 1.
    \item $c_1(a_2 | a_1^*)$ is agent 1's conjecture as to what action agent 2 will take.
\end{itemize}

The authors include additional technical details for when actions are simultaneous. When actions are determinate and outcomes visible, trust boils down just to conjectures $c$. Most centrally, however, this model defines trust as the product $$F(\cdot)\cdot c(\cdot)$$ of the outcomes function and the conjecture function across different possible worlds, weighted by the probabilities of those worlds. It is Bhattacharya et al.'s opinion that economists focus too much on the former term, while psychologists focus too much on the latter.

The paper also provides a non-mathematical definition for trust:
\begin{displayquote}
    ``Trust is an expectancy of positive (or nonnegative) outcomes that one can receive based on the expected action of another party in an interaction characterized by uncertainty.''\autocite[462]{bhattacharya}
\end{displayquote}

As stated, this theory is one of so-called ``expected outcome'' trust. Agents should be more trustworthy precisely when the expectancy of a positive outcome for them is higher, with no exceptions for cases like therapeutic trust mentioned above. Also, as a price of their mathematical rigor, expected outcome models ignore non-cognitive cases of trust, such as that which occurs between infants and their caretakers, which surely are of interest to philosophers.

\subsection{Trust in Psychology}
Psychological theories of trust, predictably, emphasize trust's interpersonal aspects. Psychologist Amy Edmondson defines trust as giving another the benefit of the doubt, plus expecting behavior in line with enabling collaborative goals.\autocite{edmondson} There is also convergence in psychological literature that trust necessarily involves accepted vulnerability,\autocite{edmondson,creed-miles-trust-organizations,kramer1999} as in philosophy.

Roderick M. Kramer, in keeping with Bhattacharya et al., notes the difference between economic and psychological theories of trust. Kramer distinguishes the \textbf{rational} and \textbf{relational} models of choice in the literature on trust. The former dominates economic theory, and as we saw above, dictates that agents assign trust exactly in accordance with their best knowledge of predicted outcomes. Kramer advocates the relational model for psychologists interested in the subject, opining that we should view trust ``not only as a calculative orientation toward risk, but also a social orientation toward other people and towards society as a whole.''\autocite[573]{kramer1999} The relational model of trust captures that in evaluating trust, human actors (not econs) are likely to make ``affective and intuitive'' rather than ``calculative'' choices.\autocite[6]{edmondson}

\subsection{Trust in Artificial Intelligence}\label{AI}
\begin{displayquote}
``Cooperation between agents, in this case algorithms and humans, depends on trust. If humans are to accept algorithmic prescriptions, they need to trust them.''
\end{displayquote}
This is a quote from the Wikipedia entry for ``Explainable artificial intelligence,'' illustrating a propensity for ``trust'' to show up in artificial intelligence literature. \autocite{xai-wikipedia} Plenty of researchers in the field of Explainable Artificial Intelligence (henceforth XAI) take trust to be a primary end goal for their research. As AI systems become increasingly complex and increasingly adopted in society, it is important that we learn to develop and evaluate trusting relationships with these systems. Trust is one dominant attitude determining the extent, productivity, and quality of our interactions with artificially intelligent systems. Moreover, AI systems are frequently adopted in modern healthcare, finance, security, military, and other contexts. A future in which trust is absent in human-machine relationships sounds increasingly dim as machines come to govern more and more of these important domains of our lives, especially in ways beyond our capacity to control or understand.

However, the operative terminology in XAI does not include trust, but instead concepts like explainability, transparency, fairness, and robustness. Instead of reaching for our goal of trust directly, XAI researchers take these other four qualities as (strongly) \textit{entailing} or (weakly) \textit{prompting} trust, or trustworthy qualities, in some way. Particularly the first two concepts, explainability and transparency, are taken to be diagnostic criteria or ``intermediate goals" for trust.\autocite{dosilovic} Researchers like Ribiero et al. have directly linked explainability to trust:
\begin{displayquote}
``Whether humans are directly using machine learning classifiers as tools, or are deploying models within other products, a vital concern remains: \textit{if the users do not trust a model or a prediction, they will not use it}.''\autocite[1]{ribiero-2016}

Emphasis in the original.
\end{displayquote}

Moreover, other researchers\autocite{pu-chen,lipton-mythos,kim_2015} have directly tied transparency to trust:
\begin{displayquote}
``Communication with the user is essential to a successful interactive system - it improves \textit{transparency}, which has been identified as a major factor in establishing user \textit{trust} in adaptive agents...''\autocite[102]{kim_2015}

Emphasis mine.
\end{displayquote}

According to Dosilovic et al., XAI researchers turn to explanation and transparency as diagnostic criteria because trust is ``hard to formalize and quantify.''\autocite[210]{dosilovic} This difficulty is worrisome -- if we do not know what trust \textit{is}, how can we take seriously any claims that explainability and transparency guarantee it? I believe modern philosophical theories on trust could provide utility, and will revisit this in the case study in section \ref{CaseStudyXAI}.

\subsection{Trust in Philosophy}\label{TrustInPhilosophy}
Of course, philosophy is where we find the greatest depth of relevant literature. Mentions of trust in philosophy date back to antiquity. In Plato's \textit{Republic}, citizens must intuitively trust their philosopher kings to rule in a good society. Yet trust, or trustworthiness, is not among the list of virtues that Plato enumerates for such a society. Annette Baier believes that the classic virtues of justice and friendship should imply one of trust, albeit indirectly.\autocite[232]{baier_1986} In medieval philosophy, Saint Thomas Aquinas wrote at length on the nature of faith, and related psychological attitude trust by proximity. Modern philosophers John Locke and Thomas Hobbes discussed trust in government and social institutions in their theories of contracts. Yet the first philosophy to really begin considering trust as a standalone notion appears in feminist thought in the 1980s.

\subsubsection{Will-Based Theories}
Will-based theories are the first family of theories we will examine. To the question of what makes trust a distinctive psychological state, these theories generally answer that trust is reliance plus some expectation of \textbf{goodwill} from the trustee, where goodwill is understood broadly as helpful, cooperative, and invested attitudes and feelings towards another person.

\paragraph{Annette Baier}\label{Baier}
Most contemporary philosophical discussions on trust begin with Baier's ``Trust and Antitrust.'' Baier's work on trust is one of the most referenced works in this subject into the present day. In this paper, Annette Baier sets out to distinguish the different forms of trust that may exist and try to formulate a \textit{moral basis} for trust -- practically speaking, a test to check whether a given instance of trust is moral.

Writing in 1986, Baier notes a ``silence'' on the topic of trust within moral philosophy. In her view, philosophers of the day had been overly concerned with the types of interpersonal relationships best characterized by moral \textit{contract}; that is, equal-footing relationships between autonomous individuals, mostly men. In Baier's view, this fixation causes us to ignore most of the instances where trust is relevant, namely in relationships of imbalanced power. She looks particularly at trust in infant-parent relationships, as well as patriarchal husband-wife relationships.

In general, Baier thinks that trust is nearly always present, even in fleeting interpersonal interactions, and we only really notice it when it suddenly goes missing. She distinguishes trust from ``mere reliance'' by noting that when trust is lost, we are liable to feel \textit{betrayal}.

Because of this point about betrayal, Baier concludes that trust is reliance with an expectation of \textit{goodwill}. Baier sees trust predominantly as a three-place predicate, where $A$ trusts $B$ \textit{with} valued item $D$, and also predominantly as a relation between two individuals.

Baier's account is influential and worth understanding, given that the majority of contemporary philosophers working on trust will cite this paper at some stage in their literature review. Yet the viewpoint has not escaped criticism. Note that Baier's tripartite form is nonstandard -- $A$ trusts $B$ with a valued item $D$, not to perform some action $X$. Such a view paints trust as \textit{entrusting} something to someone. This form works great for, say, trusting a babysitter with the well-being of a child. Yet Margaret Urban Walker notes a general difficulty when we think about, say, trusting our friends to be honest or judicious. What could we possibly say we are \textit{entrusting} them with? We really trust friends \textit{to be honest}, that is, to behave in accordance to a norm. It seems a point against the entrusting model that it cannot capture common notions of trust like this one.

Richard Holton also attacks the sufficiency of this will-based theory.\autocite[65]{Holton1994-HOLDTT} There are instances where one relies on the goodwill of another without trusting them. A fraudster uses just such a concoction to gain access to someone's bank account. In this case, the fraudster \textit{relies} on their victim to, say, provide a password, and also \textit{expects goodwill} from the victim, who presumably believes they are being assisted in regaining access to their account. Yet the fraudster does not trust their victim at all. This is often called the ``confidence trickster'' objection.

While objections seem to undermine the credibility of Baier's theory, her work is celebrated and widely cited in philosophy as well as bioethics and law as one of the first comprehensive treatments of trust in recent memory. Baier's work is insightful in articulating the reasons we trust, and her ideas inspire the conditions for the Contextual Trust Account I introduce in section \ref{ContextualTrust}.

\paragraph{Karen Jones}
Karen Jones extends Baier's original goodwill account of trust. To Jones, trust is an attitude of optimism about a trustee's goodwill and competence in the domain of interaction, such that the trustee is moved ``directly and favorably'' by the thought they are being trusted. Note that Jones has preserved the notion of goodwill in the theory while moving back to the standard tripartite model, where $A$ trusts $B$ to do $X$. Jones is thus spared objections about Baier's strange ``entrusting'' language, though other issues arise with her theory. Jones herself, in later publications, worries about the scope of goodwill as it applies in this view. Construing goodwill too narrowly seems to limit trust to the domain of only longstanding interpersonal relationships, and clearly misses some important cases, like trusting strangers. Yet construing goodwill broadly carries the risk of turning it into some ``meaningless catchall'' that just ``reports the presence of some positive motive,'' rather than picking out anything distinctive of the truster or even anything directed towards the trustee at all.\autocite[67]{jones2012}

Margaret Urban Walker also takes issue with Jones' requirement that trustees be moved ``directly and favorably'' by the thought that they are being trusted. It seems we often trust strangers, say, to respect our walking space on the street, without requiring their direct and favorable attitude towards us. Other authors working in such applied settings as bioethics, like Orona O'Neill, have agreed that Jones' direct and favorable condition leaves out many cases of genuinely interesting trust.\autocite{oneill_2002}

\subsubsection{Trust-Responsive Theories}
Following will-based accounts in the literature are ``trust-responsive theories,'' which generally take trust to be reliance plus a readiness to respond to one's own trusting in distinctive ways. So, if I trust you to drive my car, I rely on you to drive it properly. But I also dispose myself to react negatively to failures of that trust, such as your crashing my car or returning it devoid of fuel. Since this readiness to respond applies to the truster, not the trustee, these views avoid the type of objection levied above at Jones' will-based view.

\paragraph{Richard Holton}
Philosopher Richard Holton's 1994 paper ``Deciding to Trust, Coming to Believe'' is an important, and contrasting, parallel to Baier's work. Holton uses the example of the trust fall, familiar to us from drama class, as a central illustration. To Holton there are moments where we \textit{decide} to trust. The tipping point at the top of a trust fall is one such moment.

Holton's observation is that trust need not require belief -- I can trust fall without believing my drama class partner will catch me. Something other than belief makes trust a distinctive state of mind. To elucidate, Holton applies what P. F. Strawson in ``Freedom and Resentment'' called the \textit{reactive attitudes}.\autocite{Strawson1962-STRFAR} When we trust someone, so says Holton, we are prepared to react to that trust in particular ways. Should they let down our trust, we are liable to feel betrayal or resentment. Should they assist us, we are liable to feel grateful. In short, trust makes fitting the type of attitude Strawson calls a \textit{participant attitude}, and which Holton calls a \textit{participant stance}. For Strawson, the participant attitude is one we take up towards another when we are ready to hold them responsible. The participant attitude, and thus the participant stance, is loaded with normative expectations about another's behavior. This is why we react sharply to instances of betrayed trust, and are merely disappointed when our reliance in someone or something goes foul. When our trust is betrayed, we were not just expecting that someone would \textit{do} something, in the thin, rationalist sense we saw in economic theories -- we expected it \textit{of} that person. In Walker's words, on such a view ``trust links reliance with responsibility.''\autocite[80]{walker_2006}

\paragraph{Margaret Urban Walker}\label{Walker}
Walker's own conception of trust, the one she uses in her book \textit{Moral Repair}, borrows heavily from Richard Holton and generally the trust-responsive theories. Walker proposes we think of interpersonal trust as
\begin{displayquote}
    ``generically as a kind of reliance on others whom we expect (perhaps only implicitly or unreflectively) to behave as relied upon (e.g., in specified ways, in ways that fulfill an assumed standard, or in ways so as to achieve relied-upon outcomes) and to behave that way in the awareness (if only implicit or unreflective) that they are liable to be held responsible for failing to do so or to make reasonable efforts to do so.''\autocite[80]{walker_2006}
\end{displayquote}

Walker's view constrains trustees to behaving with merely ``implicit and unreflective'' awareness that they are trusted. This constraint helps her view avoid her own objection to Karen Jones. To Walker, trust is reliance plus responsibility of a weaker kind than what Jones requires. Passersby on the street can fulfill this kind of responsibility by behaving in the ways appropriate to their station, and they need not do so while explicitly evaluating you as trusting them. This construction helps Walker account for our trust in the varieties of people, ``seen and unseen,'' in our everyday life.

Beyond her theory for interpersonal trust, though, Walker's work goes on to tackle a distinct phenomenon, that of \textit{default trust}. Default trust is, at best, indirectly interpersonal, and is characteristic of the way we seem to trust businesses or institutions. When we go about engaging with large corporate entities -- say I am booking airline tickets -- we come to expect a standard for adequate performance. There is a cast of individuals behind the scenes of every major airline, and their coordinated efforts will determine whether the service I encounter is up to standard. Walker thinks, in this way, that we can trust ``the reliable good order and safety of an environment.''\autocite[84]{walker_2006}

Suppose in my airline example I am able to purchase my ticket, pass airport security, and arrive at my gate, but am unexpectedly booted from the flight after the airline discovers they've double-booked my seat. The outrage I am liable to feel at this oversight feels stronger than mere disappointment from misplaced reliance. In fact, it feels precisely like the kind of participant attitude comprising Holton's participant stance. I \textit{trusted} the airline -- in this case, some causal chain of ticketing agents, software engineers, and others unknown to me -- and they betrayed my trust. Walker says that my default trust extends to everyone in the causal chain between me and my now useless airline ticket. I extend my reliance to, and respect responsibility from, people I have never met and never will meet. It even seems that my default trust, in Walker's language, ``diffuses'' across the airline in a particular way. When our default trust is diffuse,
\begin{displayquote}
    ``we do not rely on $X$ to do $A$ and $Y$ to do $B$\dots but rather\dots we expect `reliable, courteous, and orderly service' of the airline, which is not just a particular group of unnamed individuals but a mode of organization that is supposed to train and enable whatever individuals are filling organizational roles to perform effectively to the end we rely on.''\autocite[85]{walker_2006}
\end{displayquote}
We have such default trust in more positive cases, as well, such as the communities in which we can feel respected and safe. Walking down the street in our home neighborhoods, if we are fortunate, we can feel default trust toward strangers to leave us be, or to notify us if we've dropped our wallets or keys. These are ``zones'' of default trust according to Walker. Diffuse default trust in such zones can extend to organizational entities, such as the good functioning of our law enforcement apparatus, train schedules, and the like. Default trust can embed us intimately into an environment, and can explain the high level of interdependency that many modern individuals learn to enjoy.

In section \ref{Directedness} on the settings in which trust can apply, I will object to Walker's condition that zones of default trust can only involve human actors as trustees. Despite this shortcoming I think the theory has many merits, and has been well-regarded in the philosophical literature since its publication.

\subsubsection{Trust in the Philosophy of Science}
Trust also has an active role to play in the philosophy of science, and even scientific discourse itself, so argue certain philosophers. John Hardwig's 1991 paper, ``The Role of Trust in Knowledge,'' seeks to challenge a conventional notion in epistemology that ``knowledge rests of evidence, not trust."\autocite[693]{Hardwig1991-HARTRO-3} Instead, Hardwig asserts that as modern knowledge acquisition increasingly comes to rely on teamwork and cooperation, it is our trust in others, and not our independent evidence, that serves as the foundation for our knowledge.

This is an uncomfortable account because, according to Hardwig, trust is  ``blind.''\autocite[693]{Hardwig1991-HARTRO-3} Yet in many settings it is the only way to acquire knowledge, particularly in modern science. Hardwig references a 1983 paper measuring the lifespan of charm particles in particle accelerators.\autocite{PhysRevLett.51.156} Meta-analysis showed the experiment took a total of 280 person-years of work to complete, summing across each contributor's commitments. No individual human would be capable of coming to the experiment's conclusions on their own. So, says Hardwig, with any conventional epistemological account it would be difficult to say that anyone \textit{knows} the lifespan of charm particles. But this is uncomfortable, as we want the charm particle experiment to count as useful science -- to count as knowledge. According to Hardwig, we need to grant the team itself sufficient evidence to justify the conclusion about charm particle lifespan, even though no individual team member may possess this evidence.

Hardwig states that trust factors into the origins of someone's knowledge, as well as the context of their justification. He develops the \textbf{principle of testimony} to explain cases of knowledge transfer with trust:
\begin{displayquote}
If $A$ has good reasons to believe that $B$ has good reasons to believe $p$, then $A$ has good reasons to believe $p$.
\end{displayquote}

We may also replace ``has good reasons to believe" here with ``knows" for a stronger version of the principle.

In this way Hardwig believes that we can acquire belief through testimonial evidence. Sometimes, so claims Hardwig, the best reasons for justifying a belief will be testimonial, as in the case with charm particles, simply because good reasons resting on direct, non-testimonial evidence would be impossible to obtain. Hardwig highlights the ``blindness" of this kind of knowledge: the reasons for justifying $p$, and $A$'s belief that $p$, are reasons that $A$ herself does not have.\autocite[699]{Hardwig1991-HARTRO-3}

At this point, Hardwig argues for the following disjunctive conclusion:
\begin{enumerate}
    \item There can no longer be knowledge in a lot of scientific communities relying on cooperation;
    \item One can know $p$ without having access to some of the best evidence for $p$; \textit{or}
    \item Some knowledge is actually known by \textit{teams} and not individual people on those teams.
\end{enumerate}

Hardwig decides to argue for a version of 2, which says that $A$ can know $p$ without direct access to the best evidence. This claim requires a modification to our account of rational belief.

Following this exploration into belief obtained via testimony, Hardwig moves to investigating the properties of $A$'s relationship to $B$ that might imply ``good reasons" to believe $B$'s account that $p$.\autocite[700]{Hardwig1991-HARTRO-3} These qualities of $B$ are fourfold:
\begin{enumerate}
    \item \textbf{truthful}: $B$ is being honest.
    \item \textbf{competent}: $B$ knows what generally constitutes good reasons to believe in this domain.
    \item \textbf{conscientious}: $B$ has done their work carefully.
    \item \textbf{adequate epistemic self-assessment}: $B$ must not be misled about the limits of their own knowledge in the subject matter pertaining to $p$.
\end{enumerate}

So, $A$ makes an assessment of $B$'s character, both her \textit{moral} (truthfulness) and \textit{epistemic} (competence, conscientiousness, adequate self-assessment) character. So, finally, $A$ trusts $B$ insofar as they are relying on $B$ and expecting a responsive form of goodwill or virtue from $B$ to cement that trust. Hardwig's account concerns trust in a distinct epistemic setting, yet still aligns strongly with will-based theories like Annette Baier's and Karen Jones'.

\subsubsection{Impersonal Theories of Trust}
Both Walker and Hardwig's accounts of trust, in their discussion on group settings, suggest a flavor of a different kind of trust, one not quite so wedded to the interpersonal setting. Especially in the last two decades, theories of ``impersonal'' trust in philosophy have enjoyed traction. \textbf{Impersonal trust} theories attribute trust to non-human agents, or more radically, plain objects. Entities like corporations, judicial bodies, and algorithms can be trustees under these theories. In what follows, we survey some of the better-known attempts at this philosophical tack.

\paragraph{Jeff Beuchner and Herman Tavani's ``Trust and multi-agent systems''}
In their 2011 paper, philosophers Jeff Buechner and Herman T. Tavani defend and extend Margaret Urban Walker's conceptions of zones of default and diffuse default trust.\autocite{buechner-tavani} Buechner and Tavani argue that experiments on commitment and trust within multi-artificial-agent systems should be instructive to ethicists under this diffuse default model of trust -- in other words, that philosophers can learn from experiments involving artificial agents as well as genuine ones.

According to Buechner and Tavani, the following conditions pertain to a trusting relationship between $A$ and $B$:
\begin{enumerate}
    \item $A$ has a \textit{normative expectation} that $B$ will do such-and-such;
    \item $B$ is responsible for what it is that $A$ normatively expects her to do;
    \item $A$ has the disposition to normatively expect that $B$ will do such-and-such responsibly;
    \item $A$'s normative expectation that $B$ will do such-and-such can be mistaken;
    \item Subsequent to the satisfaction of the above conditions, $A$ develops a disposition to trust $B$.
\end{enumerate}

The authors call these conditions the ``Strawson-Holton-Walker" model of trust, relying on contributions from philosophers P. F. Strawson, Richard Holton, and Margaret Urban Walker.\autocite{Strawson1962-STRFAR,Holton1994-HOLDTT,walker_2006} Buechner and Tavani argue that notions of ``normative expectation" and ``responsibility" make this theory difficult to extend to cases involving artificial agents, to which we cannot simply ascribe responsibility. They incorporate another view, also from Walker, in which artificial agents are ``part of the environment," and people come into trusting relationships with that environment. 

Margaret Urban Walker's account of trust, as we have just seen in section \ref{Walker}, begins with the observation that trust is contextual and localized in space; in particular, there are places we feel safe and operate with varying minimum ``default" levels of trust. Walker's examples often entertain large communities where such ``trust zones" occur, like entire cities where we expect adherence to traffic laws from one another, including pedestrian foot traffic ``laws" that aren't enforced (e.g., making way on the subway platform). We take such actions to be the responsibility of people within the community, and it seems that complete strangers can have these symmetric and reversible normative expectations of each other. Walker considers this form of trust an ``unreflective and habitual background" in many scenarios we find ourselves in.

Because zones of default trust may contain people who rarely or never meet in person, and yet nonetheless trust one another, the explicit individuals in the trust relation need not be spelled out in advance. Instead, Buechner and Tavani propose the notion of a ``generic individual" partaking in the zone of default trust, which may be surrogate for a real individual, a group, or (here diverging from Walker) a non-human agent like a computer network.\autocite[43]{buechner-tavani} It is this final observation that makes the Walker-Buechner-Tavani model a model for impersonal trust.

Walker, herself, considers a variation of this idea called \textit{diffuse} default trust, as we explored above. Her categorical example involves feeling resentful towards an airline for poor service, say after a day full of delayed and cancelled flights. When we feel let down by an entire airline, says Walker, it's not that we were relying $X$ to do $A$, $Y$ to do $B$, etc., but rather the airline itself, which is a ``mode of organization that is supposed to... enable whatever individuals are filling organizational roles."\autocite[85]{walker_2006} To Buechner and Tavani, it's unacceptable to reduce this down to trust of each airline employee, since we very well say and mean that we trust things like airlines, and this deserves a semantic account distinct from the state that would be trusting each airline employee. The upshot here is that the organizational or operational mode that is the airline is a recipient of trust under Walker's diffuse, default model, and Buechner and Tavani make clear that things like networks of artificial persons can likewise be recipients of trust in this way.

It is worth noting that neither Walker nor Buechner and Tavani apply the diffuse default model explicitly to the case of explaining or trusting technology. Buechner and Tavani's intentions for promoting the view have to do with refining our ethical account of trust in interpersonal relationships, aided by experiments involving artificial agents. The fact that such artificial agents can be spoken of as trusting and trustworthy under this view, however, is what makes the model relevant for our discussion in later sections.

\paragraph{``E-Trust''}\label{ETrust}
Another theory of impersonal trust gives additional focus to artificial agents, allowing even more flexibility in its application. Philosopher Mariarosaria Taddeo, together with logician Giuseppe Primiero, develops a model of a distinct notion, ``e-trust," which is trust occurring in online or digital environments.\autocite{taddeo_2010,PRIMIERO201292} Specifically, the model applies to interactions between Artificial Agents (AAs), which allows the decision calculus to be fully rational.

The authors summarize e-trust with the following four claims:
\begin{enumerate}
    \item E-trust is \textit{rational}, specifically appealing to Kant's regulative ideal of a rational agent, in which the agent chooses the best option for itself given a specific scenario and goal-orientation. This is another rationalist account of trust, in Roderick Kramer's rational / relational terms for the breakdown of trust in psychological theory, though Kant's regulative ideal introduces nuance missing from our examples from economics.
    \item From the above, e-trust is both goal-oriented and action specific. In other words, it is permissible to trust an AA at one task but distrust them at another task; e-trust is not a global property given to AAs.
    \item E-trust is a second-order relation that affects first order relations characterizing actions. For example, if AAs $A$ and $B$ transact via the sale ($S$) of some good ($g$), then $S(A,B,g)$. E-trust, $T$, is a second-order relation over transactions like $S$, meaning $T(S(A,B,g))$ will affect the conditions under which $A$ sells $g$ to $B$.
    \item E-trust has the property of minimizing the truster's effort and commitment to the achievement of a given goal. This happens by delegation of an action to the trustee, together with limited supervision of the trustee. The less a truster trusts, the more they will supervise, or even replace, the actions of the trustee.
\end{enumerate}

Roughly, an algorithm for assessing trustworthiness between AAs is spelled out like the following: an AA calculates the ratio of successful actions to total actions performed by the potential trustee to achieve the same or similar goals.\autocite[7]{taddeo_2010} Under this algorithm, e-trust is not calculable \textit{a priori}, since the truster needs previous actions from the trustee in order to assess it.

Lastly, Taddeo indicates that extending the work to more complex cases, such as those where human agents (HAs) are either trusters or trustees, would be more complex. These cases bring attitudinal and psychological factors into play, where previously only economic factors (rational factors) were relevant. It is for this reason that Taddeo dubs her concept ``e-trust'' and not actually trust.

\paragraph{C. Thi Nguyen's ``Unquestioning Attitude''}\label{Nguyen}
In ``Trust as an Unquestioning Attitude," C. Thi Nguyen establishes a theory of trust to accompany philosophy's typically agent-oriented theories. Most previous theories of trust hold only between agents, involve central requirements like goodwill, responsiveness, or reliance on commitments, and are marked by the possibility of betrayal.

Nguyen, by contrast, establishes a theory of trust based on an ``unquestioning attitude'' -- by this theory, to trust something is to have a strong disposition to suspend deliberation about that thing.  Nguyen defines: to ``trust $X$ to $P$" is to:
\begin{enumerate}
    \item be first-order disposed to immediately accept that $X$ will $P$, and
    \item to be second-order disposed to deflect questions about whether $X$ will $P$.
\end{enumerate}

This two-tiered dispositional formula mirrors Michael Bratman's two-tiered account of resolutions.\autocite{Bratman1987-BRAIPA} For example, a climber trusts their climbing rope (to hold them) because they aren't constantly reassuring themselves as to the rope's integrity. By contrast, to \textit{distrust} is to remain in a state of constant questioning and skepticism.

Nguyen claims that both theories of trust -- the general agent-oriented theory and the unquestioning attitude theory -- sit under an umbrella of a more general notion of trust, and that what joins them is their purpose of expanding one's agency through integrating aspects of the external world. There is too much information in the world for one agent to account for all at once -- as we covered with ``information overload'' in section \ref{ValueOfTrust} on the value of trust. Therefore, we are required to trust and hold the unquestioning attitude towards certain things. In this sense, we trust so to form linkages to external objects, and we bring them into our practical and cognitive faculties. For this reason, we can talk of being \textit{betrayed} in our trust even of non-agents, like musical instruments or the ground, because the ``normative bite" comes from our desire to integrate objects in our immediate cognition and agency. When our climbing rope unexpectedly snaps, we feel betrayed because we had integrated that object into our agency and had developed an attitude of not questioning its integrity.

While initially attractive, this construal presents difficulties. Suppose I am aide to some despotic ruler, whom I despise. The ruler informs me of his plans to invade a neighboring sovereign nation. Convinced of his bellicose nature, I am first-order disposed to immediately accept that he will invade the nation. Also, fearful of my ruler's wrath, I am second-order disposed to deflect questions about whether he will invade. Such a disposition could apply not just to explicit verbal questions, but to my own internal -- even subconscious -- questioning. Perhaps I endured some torturous initiation that permanently erased any tendency for me to question him. Is it right to say I \textit{trust} the ruler to invade the nation?

Nguyen's view says I do trust the ruler to invade, and this result feels incorrect. I certainly \textit{expect} him to invade. I \textit{believe} he will invade. Yet I also dread his coming invasion, since I despite him. It feels incorrect that one would both trust and dread the same action. The view seems to be missing something about the affect that should accompany trust. Recall that will-based theories like Baier's or Jones' set out to distinguish trust from reliance. One essential feature signifying the difference, among several, would be a positive inclination towards the action one is trusting. Likewise, a trust-responsiveness account like Holton's would likely include a normative basis for one's reason to trust, couched in something like an investment in the action's outcome. Nguyen's account does not preserve these features, and I think some of the resulting cases counting as trust are peculiar. As a matter of psychology, we \textit{want} to trust. Yet, the aide would certainly prefer to be wrong about the coming invasion. Further complicating the picture is the fact that a second-order disposition to deflect questions can come about via coercion, not just via certainty. The very phrase ``deflect questions'' brings to mind a press secretary fumbling through an interview after a particularly objectionable policy has been passed.

All together, this objection targets the sufficiency of Nguyen's account. An unquestioning attitude may still be necessary for trust, and indeed I will argue this case in section \ref{DefiningContextualTrust}.

\section{Difference-Making Features of Trust}\label{DifferenceMakingFeatures}
Thus concludes our literature review of trust in recent philosophical history. It remains to put this lengthy literature review to use, and attempt to distill some commonalities.

In thinking about the concept of trust as articulated throughout section \ref{Background}, three features come to mind. First, trust is a relation between two individuals. You cannot trust, except in someone or something, and you cannot be trusted, except by someone or something. Call this type of trust \textbf{directed} trust. Philosophers seem to agree that trust is generally directed.\autocite{baier_1986,sep-trust} Second, you can trust someone to perform certain actions, yet not others. I trust my plumber to fix my sink, though not necessarily to pet-sit my dog, or to get to my house without directions. Call this type of trust \textbf{action-sensitive} trust. Philosophers seem to be in agreement that trust is generally action-sensitive.\autocite{baier_1986,hardin_2004,Nguyen-trust} Third, you can trust someone to perform an action in some contexts, yet not in others. I trust my plumber to fix my sink, though not necessarily while juggling, or while the sink is on fire. Call this type of trust \textbf{context-sensitive} trust. If philosophers are in agreement that trust is context-sensitive, they have not said so (as we saw in the literature review). Russell Hardin writes that ``trust is generally a three-part relation: $A$ trusts $B$ to do $X$.''\autocite[9]{hardin_2004} Such a construal captures directness and action-sensitivity, but not context-sensitivity. I will argue this omission is a mistake.

It seems clear that when we use the word ``trust,'' we take context into account. ``Do you trust your friend to drive that car?'' ``Not when it's raining like this!'' Even in cases like ``Of course I trust my spouse with the baby!'' context appears relevant. Either context is implicitly understood -- e.g., the listener knows the spouse is just putting the baby to bed -- or deliberately omitted to strengthen the statement -- e.g., the speaker asserts that they trust their spouse with the baby \textit{no matter what happens}. In either case, context plays a role in determining exactly what kind of trust is present.

Of course, one could object that this feature of language is not at all unique to trust. This is so: in stating that ``I like ice cream \textit{when it is hot outside},'' the contextual factor ``hot outside'' plays a role in determining exactly how I like ice cream. This observation is not groundbreaking in the case of liking ice cream, and does not seem any more so in the case of trust. Maybe the omission of context-sensitivity from prior philosophy is warranted for this reason -- maybe context-sensitivity is mundane, and so assumed in the background? I will have more to say on this objection in section \ref{MundaneContextObjection}. For now, two points suffice.
\begin{enumerate}
    \item If we intend to take seriously the idea that trust is a ``three-place relation,'' then we ought to consider seriously what a \textit{relation} is. A relation characterizes a property that holds between certain groups of objects and not others. So, a three-place relation holds between certain triples of objects and not other triples. If trust is a three-place relation like this, then trust either holds for $\langle\text{`me'}, \text{`you'}, \text{`driving my car'}\rangle$ or it does not. Now, if I only trust you to drive my car when it's not raining, then trust does not simply hold or fail to hold for $\langle\text{`me'}, \text{`you'}, \text{`driving my car'}\rangle$. Instead the relation ``trusts-when-it's-not-raining'' would be the closest thing, since it simply holds for that given triple. If one now objects to say that trust cannot be the type of thing that simply holds like this, then trust cannot be a \textit{relation} in any precise sense.
    \item As we will see in section \ref{ContextSensitivity} on context-sensitivity, not just anything gets to count as a contextual factor. In this way, we don't get to say ``$A$ trusts $B$ to do $X$ if $\Delta$,'' where $\Delta$ is some arbitrary conjunction of facts. A context, $C$, will turn out to be a structured set of facts about the truster $A$'s beliefs and past observances of $B$ doing $X$. In this way, context-sensitivity can be informative about trust, and does not just catch all edge cases in some convenient but ultimately vacuous way.
\end{enumerate}

Once again, I argue that any kind of trust worth investigating must be directed, action-sensitive, and context-sensitive. So, rather than answer the question ``What does it mean to trust?,'' I propose instead to tackle the more specific ``What does it mean for $A$ to trust $B$ to do $X$ in context $C$?''

Once we have established these three properties as relevant for trust, we are in a position to talk about the nature of trust.

I will argue later that $A$ trusts $B$ to do $X$ in context $C$ just in case $A$ takes $B$'s doing $X$ as a means to one of $A$'s ends, \textit{and} subject to $C$, $A$ adopts an unquestioning attitude towards $B$'s ability to $X$. The former condition takes inspiration from Richard Holton,\autocite{Holton1994-HOLDTT} Karen Jones,\autocite{jones-affective-attitude} and Margaret Urban Walker,\autocite{walker_2006} and the latter from C. Thi Nguyen\autocite{Nguyen-trust}, though no one of these views fully encapsulates both conditions. Before this argument, though, we must make the case for directed trust, action-sensitive trust, and context-sensitive trust individually.

\subsection{Directedness}\label{Directedness}
Trust is \textbf{directed} if $A$ can trust $B$ but also not trust another trustee $B'$, and also if $B$ can be not trusted by some other truster $A'$. There is no disagreement in philosophy that trust has a directed character. Trust is a transitive verb, hence phrases like ``Alice trusts,'' or ``Bob is trusted'' are elliptical: Alice must trust some $X$ (or collection of $X$'s) and Bob must be trusted by $X$ (or collection of $X$'s). So, trust is \textit{at least} a two-place notion, such that $A$ trusts $B$.

Philosophers like to use this phrase ``$n$-place notion'' or ``$n$-place relation'' with respect to trust, as in Jones:
\begin{displayquote}
    ``There is also general agreement that trust is a three-place relation: $A$ trusts $B$ to do $Z$.''\autocite[668]{Jones2015}
\end{displayquote}
I interpret this phraseology as identifying the number of dimensions along which trust can \textit{vary}. I have not seen this interpretation in other literature, but I think it helps to specify the question of the nature of trust.

Consider the two-place case first. When $A$ trusts $B$, $B$ is \textbf{difference-making} in the sense that, were you to replace $B$ with some $B'$, $A$ might not necessarily trust $B'$. Likewise, $A$ is difference-making since some other $A'$ might not trust $B$, while $A$ does.

Thus, the debate in contemporary philosophy is not as to whether $A$ and $B$ are difference-making; rather, philosophers disagree on what gets to be an $A$ or a $B$.

Take first the case of $A$, the truster. In our literature review, Mariarosaria Taddeo and Giuseppe Primiero (section \ref{ETrust}) were the only authors explicitly allowing for $A$s to be ``Artificial Agents,'' meaning non-humans. Tellingly, these authors do not even call their subject matter ``trust,'' instead opting for ``e-trust'' to designate trust occurring in online or digital environments. Here, there is explicit indication that the phenomenon being explored is different than ``full-blooded'' trust (to borrow a phrase from Baier). We might also see a signal for this in the fact that e-trust is entirely rational. Most philosophers and psychologists feel that a completely rationalist account of trust fails to properly characterize the state of the truster -- some indication of relational decision-making is needed.

Other philosophers, whether explicitly or by implication of their theory, leave out the possibility that non-human agents can play the role of $A$. Baier mentions explicitly that her account is intended just for interpersonal contexts. Jones, in attributing an attitude of optimism about the goodwill of the trustee, makes it difficult to imagine how a non-human agent could trust under her definition. Also, Walker explicitly restricts her trust-responsive theory to interpersonal cases, and Richard Holton's participant stance seems only possible if $A$ can have human-like intentions and mental attitudes. So, keeping with the grain, we ought to restrict $A$ to just human agents.

Since human agents play the role of trusters, and since we reject a thoroughly rationalist account of trust for reasons of insufficiency, we can make a further point about the role of $A$. Say directed trust is \textbf{truster-subjective} if $A$ gets to be the ultimate arbiter of whether they trust $B$. In other words, no objective account of $A$'s mental state can override $A$'s decision to either trust or not trust $B$. On its face this assumption feels unproblematic. Note that this does \textit{not} entail that $A$ has willful control over the cases in which they trust, just that when they \textit{do} trust or not, their state cannot be somehow incorrect. In real cases where people divulge their trust, or lack thereof, in some $B$, we never really have grounds to correct them. Such a notion does not really make sense.

Now for $B$, the trustee, where my opinion may diverge slightly more from the philosophical mainstream. Up to this point, I have liberally employed phrases like ``trusting an $X$,'' with $X$ substituted for an object, in several sections. We trust airlines to deliver valid tickets, we trust ropes to hold our weight, and we trust newsfeeds to deliver truthful content. Such phrases, I think, read quite nicely. It is natural to talk in this way. Natural, I think, because the concept of trust naturally includes objects as genuine recipients of trust.

An examination of Walker's zones of default trust, from section \ref{Walker}, will support this point. I think Walker's account with default trust is right on the money, though her insistence to restrict zones of default trust to the strictly interpersonal is puzzling. Consider the case with the airline again. In my ticket purchasing example, Walker concedes that the airline realizes a ``mode of organization'' that trains and enables airline employees to help me with my travels. Can I not trust the organization \textit{itself}? When I say I ``trust American Airlines'' it feels much more accessible to believe that my statement is directed towards that very object, American Airlines, rather than shorthand for ``I trust employee $X$ to do $A$ and $Y$ to do $B$ and\dots.'' Many philosophical views allow organizations like airlines to count as agents in this way.\autocite{floridi-sanders} This feels especially apt in the case of diffuse default trust, where I trust ``reliable, courteous, and orderly service'' of the airline itself. In cases (like airline ticket purchasing) where I do not know many of the intermediary steps, my trust in any particular middleman of the transaction feels ghostly, and not like something I could defend for myself. In addition, nowadays the \textit{thing} actually \textit{booking my ticket} is not a human at all, but a script on a web server somewhere. If booking a ticket is an action (which it surely is), and moreover an action in the causal chain of me booking travel plans, it stands to reason that my trust can be directed at computer systems in addition to airline employees. Walker is insistent that default trust only extends to the human individuals at an organization, and I cannot quite see why.

We can even construct a more direct example to get at this latter point. Suppose I am opposite a Tesla while stopped at some Bay Area intersection. The Tesla has its right blinker on. Thus, when the light turns green, should I decide I trust the Tesla driver to use his indicator, I can drive straight through the intersection without risk of collision. In a certain light, cars are ``modes of organization'' that rely on collaboration between the driver, steering column, brake pedals, and so on for effective functioning.

Now for the twist. Suppose the front windshield of the Tesla is tinted, meaning I cannot tell if a driver is at the wheel of the car, or if Tesla's Autopilot feature is driving. If the former, then when I drive through the intersection, according to Walker, I trust the human driver in the context of her ``mode of organization,'' namely the various subcomponents of the car. Yet, if the latter, then there is \textit{no} human agent in the causal chain between the Tesla's right indicator and its turning right. Note that the car designers and manufacturers do not count, since they can make no causal difference to the car's turning right in this particular instance. Thus Walker's definition for default trust, if the human-in-the-loop is so necessary, fails to hold. When I drive across the intersection, do I then trust\dots nothing? That is, does my trust fall to the level of mere reliance if it turns out no one was driving the car? This would seem problematic. I behave identically in either scenario. And the tinted front windshield is no special case, since Walker allows for this very kind of trust towards ``unseen'' agents in her example with the airline.

It would be much preferable if my concept of trust was flexible in this case. In other words, I should be able to trust the driver of the Tesla, human or not. I conclude that if Walker's default trust account is correct, $B$ should be able to include non-human actors.

\subsection{Action-Sensitivity}\label{ActionSensitivity}
Directed trust is \textbf{action-sensitive} if $A$ can trust $B$ to take action $X$, yet not trust $B$ to take a different action $X'$. In 2015, Karen Jones observed that most philosophers took trust to be a tripartite relation, consisting of truster, trustee, and action. Thus, for philosophers, trust is at least commonly action-sensitive. Specifying a particular action over which trust is given has been successful in other accounts, and provides a good example of the ``boundary condition'' behavior characteristic of difference-making features for trust. 

Annette Baier's view of trust as entrusting is related, although distinct, from action-sensitivity. For Baier, $A$ trusts $B$ \textit{with valued item} $D$, which is not the same as trusting one to take an action. In some cases the difference-making properties of $D$ might extend to plausible actions $X$. For example, if you entrust me with your french press but not your white sneakers, presumably you trust me to brew coffee but not to keep your shoes spotless. For other entrusted items, though, no direct parallel seems clear. Entrusting someone with your life or a sensitive family secret may make a wide range of action responses appropriate, depending on the content of either. In fact, this failure to present her trust as action-sensitive is precisely what Margaret Urban Walker criticizes about Baier's theory. As Walker points out, when I trust my friends to be courteous or honest, with what could I possibly be entrusting them?

A host of other theories indicate action-sensitivity much like Jones and Walker do. Nguyen points out that ``when I say I trust my doctor, I can usually be understood to mean that I trust my doctor to perform their medical duties, and not that I trust them to successfully do modal logic or play jazz.''\autocite[21]{Nguyen-trust} So, action-sensitivity in the nature of trust seems well-established.

\subsection{Context-Sensitivity}\label{ContextSensitivity}
Finally, directed, action-sensitive trust is \textbf{context-sensitive} if $A$ can trust $B$ to do $X$ in context $C$, but potentially not trust $B$ to do $X$ in differing context $C'$. We have already seen intuitive examples of this at play. I trust my friend to drive my car in the suburbs, but not the city. I trust my wool jacket to keep me warm, but not in heavy wind and rain. You trust yourself to understand this paper upon reading it, though maybe not in a nightclub or if I had written it in Klingon.

One can be tempted to blur the domains of action- and context-sensitivity, for example with confusing cases like ``I trust Margaret has studied for the test.'' Is ``studying for the test'' an action we might trust Margaret to do; or rather might we trust Margaret to \textit{study}, just perhaps not for this test? The former sets ``study for the test'' as the action; the latter sets ``study'' as the action and ``for this particular test'' as the context. I address such ambiguous cases in response to an objection in section \ref{ActionGranularity}. Of course, some cases are bound to be more linguistically ambiguous and confusing than others. However, we can get more precise with what we mean by ``context.'' There are several specific ways in which context can be relevant for trust.

First, recall that in discussing directed trust in section \ref{Directedness}, we established that trust is truster-subjective. We take this term to mean that the truster, $A$, gets the ultimate decision about whether they trust $B$ to $X$ in context $C$. Truster-subjectivity already places some limitations on the context $C$. For example, if $A$ does not know some fact $F$, $C$ cannot contain $F$ unless we are speaking counterfactually:
\begin{center}
    ``Had $A$ known $F$, then maybe they would have trusted $B$ to $X$\dots''
\end{center}
Even in these counterfactual cases, facts like $F$ only get to enter the fray when they are relevant to $A$ in some way. In fact, anything belonging to a context $C$ behaves in this way. Since $A$ ultimately decides whether to trust $B$ to $X$ (again, not necessarily willfully), any relevant contextual factors must be determined via relevance to $A$ herself. Take note that ``relevance to $A$'' does not entail the property ``known by $A$.'' If $A$ has deep-seated racial biases, for example, they will affect $A$'s ability to trust without her awareness.

This observation about truster-subjectivity imposes some structure on a context $C$. We can say that relevant factors in a context $C$ belong to one of the following groups:
\begin{enumerate}
    \item Contextual factors can be \textit{facts about $A$}, including beliefs $A$ holds about herself. For example, ``$A$ holds the following racial prejudice'' or ``$A$ has a Ph.D. in Artificial Intelligence'' can be contextual factors of this kind.
    \item Contextual factors can be \textit{beliefs $A$ has about $B$}. Note that contextual factors cannot be unrestricted facts about $B$, because facts unknown to or unconcerning $A$ will be irrelevant to whether $A$ trusts. General facts about $B$, unknown to $A$, might affect $B$'s ability to $X$, but not $A$'s ability to trust whether $B$ will $X$. For example, ``$A$ heard from her cousin that $B$ is a slimeball'' or ``$A$ and $B$ grew up in the same small town'' can be contextual factors of this kind.
    \item Also, contextual factors can be \textit{beliefs $A$ has about $X$}. This form follows a very similar form to point 2. For example, ``$X$-ing is close to impossible in the dark'' or ``I myself can do $X$'' can be contextual factors of this kind.
    \item Finally, contextual factors can be \textit{beliefs $A$ has about $B$'s history of doing $X$ in differing contexts $C'$}. This point is the most conceptually involved, since it entails that contexts $C$ have a recursive definition. Though, these kinds of beliefs are intuitive. When assessing whether to trust someone, a natural first step is to assess their track record on the action in question. A surgeon with a thousand successful surgeries under her belt should typically be more trustworthy than a novice. Track records are an important aspect of longstanding trusting relationships. Veteran football teammates have an extensive understanding of each others' intuitions and likely moves on the pitch, drawn both implicitly and explicitly from their recollections of a long shared history playing together. Likewise, an experienced soldier trusts her gun not to jam only under certain specific conditions, drawn both from her theoretical understanding of the weapon (see 2.) and a long history of past experiences. Note that the differing contexts, $C'$, need not be contexts happening earlier in time in the same possible world. This freedom will be important when we come to understanding the role of model transparency in trust in the Artificial Intelligence discipline.
\end{enumerate}
Put all together, collections of these four types comprise \textit{contexts}, and contexts can be difference-making to a case of trust between a given truster, trustee, and action. Points 2, 3, and 4 explicitly mention truster $A$'s beliefs, and so contexts can be considered similarly to belief sets in the context of Bayesian inference. Contextual trust is similar to Bayesian inference in that trust is \textit{conditioned} on some set of facts, $C$. However, we need not think of trust as anything like conditional probability, unless we adopt a rationalist account like the one discussed in section \ref{Economics}.

\section{Contextual Trust}\label{ContextualTrust}
Having explained directedness, action-sensitivity, and context-sensitivity, we have a template to which any theory of contextual trust must adhere. In what follows in section \ref{DefiningContextualTrust}, I will present my own best idea of the nature of contextual trust. Before my own approach, though, it will benefit us to review how existing philosophical theories of trust can accommodate context in the way described above. As we can see, many philosophical accounts are highly amenable to the explicit use of context.

\subsection{Contextual Trust in Previous Philosophical Theories}\label{ContextualTrustInPreviousTheories}
Treating the nature of trust contextually is not a new philosophical idea. Other theorists in the literature review above have referenced factors like contexts, though they have done so with varying and disparate terminology.

First, Margaret Urban Walker's default zones of trust,\autocite{walker_2006} and subsequent work on the idea by Buechner and Tavani,\autocite{buechner-tavani} seem to align closely with the idea of contexts. Zones and contexts are not wholly unrelated -- both involve a circumstantial backdrop in which many kinds of trust between trusters, trustees, and actions are appropriate. Moreover, this backdrop may differ for the truster in each case, even with the objective description of the zone or context held constant. However, unlike zones, contexts do not have a necessarily habitual component. Contexts can be fleeting -- consider negotiating right-of-way on a tricky roundabout in a state you'll never visit again. Also, contexts may more naturally describe scenarios in which trust is highly discouraged. A war zone, for example, is a place devoid of most zones of default trust, yet comprises a context mostly devoid of trust, which is subtly different.

Walker also mentions ``context'' in \textit{Moral Repair}, though the point is more to highlight how the \textit{motivations} to uphold trust can differ in different settings. Still speaking just of interpersonal trust, Walker says, ``we may be moved in many ways to do what we know we are responsible for doing.''\autocite[81]{walker_2006} Walker highlights how ``norms in different contexts'' can modulate our normative expectations and hence the conditions for the trust we invest in others. Depending on the circumstances, we may take evidence of goodwill, a willingness to please, or empirical guarantees of reliability as signals to trust, and each of these can be subsumed under contexts $C$. As I will articulate in section \ref{DefiningContextualTrust}, I take certain motivations to be essential to trust, and certain others to be contextual modifiers, though the construction of $C$ itself does not privilege my particular theory over any others.

Next, Nguyen's ``Unquestioning Attitude'' account of trust, explained in section \ref{Nguyen}, could be understood as involving contextual factors in an implicit way. Nguyen says that when we trust $B$ to $X$, we hold two dispositions. The first is a first-order disposition to immediately accept that $B$ will $X$. The second, the ``unquestioning'' disposition, is a second-order disposition to deflect questions about whether $B$ will $X$. We can best understand these two dispositions as characterizing the presence or absence of particular beliefs in the context $C$. When $A$ does \textit{not} trust $B$ to $X$, according to Nguyen, either or both cases are true:
\begin{enumerate}
    \item $A$ holds some first-order skepticism as to whether $B$ will $X$. Perhaps $A$ is naturally skeptical, or particularly defaults to skepticism regarding trustee $B$, or action $X$. Perhaps $B$'s ``track record'' at $X$ is too thin for $A$'s standards. Any of these reasons for the absence of a first-order disposition may be found codified in the context $C$.
    \item $A$ is not disposed to deflect at least certain questions about whether $B$ will $X$. In this case, $A$ is missing some critical information about whether $B$ will $X$, which may take the form of questions about $B$'s approach to $X$, or properties of $B$ (comfort with children, flammability, et cetera).
\end{enumerate}
By negation, when $A$ does trust $B$ to $X$, that first-order skepticism is gone, and any potentially troublesome gaps in $A$'s knowledge or expectations have been filled. Thus, to say $A$ trusts $B$ to $X$ at context $C$ is to say that the conditions of $C$ entail neither first-order skepticism on behalf of $A$, nor any gaps in $A$'s understanding that $A$ views as problematic.

\subsection{A Theory of Contextual Trust}\label{DefiningContextualTrust}
We have shown that several different philosophical theories of trust can work as theories of contextual trust. What is the best theory for contextual trust? As with other gnarled and pluralistic concepts in philosophy, there is really no ``best,'' just best suited for a particular application or domain of discourse. I will give and defend my view for a particular theory of contextual trust, though by no means is this view the ``best'' one.

We may begin with the philosophical ideas that already appear off the table. In describing Directedness, I made the case that trustees $B$ may be non-human entities. This flexibility is shared most notably in C. Thi Nguyen's view, but it immediately invalidates both Baier's and Jones' theories as theories of contextual trust. Both Baier and Jones consider goodwill to be a necessary feature of the trustee, and non-human entities like algorithms and chairs simply cannot foster goodwill towards you. Walker's opinion in \textit{Moral Repair} is also unamenable since she seems to reject cases of trust falling outside interpersonal cases (as I argued in section \ref{Walker}, without seemingly good reason). Walker seems to consider ``default trust'' to be different from trust in her writing, though it is somewhat unclear. Walker's use of ``default trust'' makes it seem possible that one might trust, say, an airline for reputable service under this definition. Thus default trust may be a way to describe contextual trust, and we already showed in section \ref{ContextualTrustInPreviousTheories} how the two theories relate.

Since Nguyen's ``Unquestioning Attitude'' account provides the most natural explanation of impersonal trust, it is a useful place to start. Yet, I already highlighted a flaw of this account in section \ref{Nguyen}, namely that the unquestioning attitude alone does not explain the often distinctive affect of trust. To his credit, Nguyen includes the reply to this objection in his paper,\autocite[6]{Nguyen-trust} just not in his precise definition of what it is to trust $B$ to $X$. We can thus slightly modify Nguyen's definition and arrive at an account that mostly matches his own.

Preamble now aside, I propose the following \textbf{Contextual Trust Account}:

Truster $A$ trusts trustee $B$ to perform action $X$ in context $C$ if the following obtains:
\begin{enumerate}
    \item Subject to context $C$, $A$ adopts an unquestioning attitude as to whether $B$ will $X$.
    \item $A$ incorporates $B$'s doing $X$ as a means to one of $A$'s ends.
\end{enumerate}
Point 1. is precisely Nguyen's unquestioning attitude, yet I define an unquestioning attitude relative to a specific context $C$. For $A$ to adopt the unquestioning attitude in context $C$, $A$ acknowledges that there is nothing about the current state of affairs that would make $A$ doubt whether $B$ can $X$ in any serious way. $A$ might certainly have counterfactual doubts about $B$ -- in another context $C'$ it may be impossible for $A$ to trust in the same way.

Point 2. encapsulates the reply to the objection to Nguyen by enforcing that $A$ is in some way \textit{invested} in a successful outcome of $B$ doing $X$. We needn't trust in cases where our trust would have no value, and so point 2. is a manner of ensuring such value. $A$ has some goal, and $B$ doing $X$ works in service of that goal. The requirements for such a goal can be very loose, such that publishing a research paper, eating a nice meal, not getting stabbed, and so on might all qualify.

Put together, points 1. and 2. tell a narrative about why we trust. As agents in the world, we have projects and goals we would like to see come into fruition. Sometimes, our individual efforts cannot suffice to actualize our goals. We can depend on others, perhaps those with talents or knowledge exceeding ours, to help in our pursuit. Yet sometimes even supervision of -- or certainty in -- the success of these others is not possible. In these cases, we may aim to trust. We adopt another agent's doing $X$ as a means to one of our ends, and via the unquestioning attitude put the details of their execution of $X$ out of our mind. Hence, trust frees our limited mental capacity to perform other tasks, including perhaps acting as trustee for another in a domain where our abilities exceed their own. In this way trust has a critical role in resource sharing and coordinated planning, and also allows us intimate positions in each others' life projects as genuine contributors.

Also, contextual trust can accommodate the increasing role that technologies like voice assistants, virtual reality platforms, and curation algorithms play in our agency and our goals. For many of us, devices like computers and smart phones are, for better or worse, indispensable companions in our social and professional lives. My computer is as intimate a collaborator in my professional success as my human coworkers, and I expect this sentiment is not unusual.

Those married to a goodwill account of trust may worry that allowing devices into the role of trustee threatens to obscure the gap between trust and reliance. Yet, the above discussion about projects shows that our connection to technological artifacts can far exceed mere reliance. Nguyen offers an agreeing commentary:
\begin{displayquote}
    ``Contemporary life is significantly marked by trust in technological artifacts and technologically-mediated social environments: Google’s search algorithms, smart phones, the ranking algorithm behind Facebook and Twitter, the emergent  networks of interconnection on social media. Our relationships with these objects, I suggest, is far more potent than mere  reliance.''\autocite[8]{Nguyen-trust}
\end{displayquote}

I should also add that such connection need not be anthropomorphic. I am under no delusions that my phone desires to help me or feels normative pressure to perform in response to my trust. Yet I do not just rely on my text messages and photos persisting between power cycles -- I genuinely \textit{trust} that they're still there, as a failure of this kind could be devastating to my ability to socialize and land a core blow to some important life projects of mine.

\subsection{Replies to Objections}\label{RepliesToObjections}
In the previous section, I established the Contextual Trust Account as one particular theory of contextual trust. In what follows we consider several objections to the arguments made thus far. I consider objections both to the central point of section \ref{ContextSensitivity}, which is that context matters in assessing trust, and that of section \ref{DefiningContextualTrust}, which presented a basic theory following the argument about context.

\subsubsection{``Implicit Context'' Objection}\label{MundaneContextObjection}
\textit{Objection}: Asserting that context ought to also be considered in cases of trust doesn't really add much. Of course previous theorists, like Baier, Jones, Walker, and Nguyen, would have acknowledged that context is relevant in assessing trust. Trust does not need to be a four-place relation, including context, though. We evaluate contextual factors when deciding to trust, though trust is still a three-part relation that holds between a truster and trustee for a given action.

\textit{Response}: To say we evaluate contextual factors when we decide to trust is, of course, true, but it does not help simplify the picture. We are not sensitive to just about any contextual factor, and some factors we only consider some of the time. Context-sensitivity, as I have defined it above, provides a way to characterize contextual factors in this way. As stated in section \ref{TrustingInContext}, context is difference-making for trust by establishing ``boundary conditions'' where trust in a given trustee-action pair is bound to fail.

Nonetheless, this objection posits that other philosophical theories of trust can succeed without mentioning context, thereby making context implicit. Take goodwill theories of trust, as from Annette Baier or Karen Jones, as examples of why this cannot work. To review, Baier says that trust is reliance plus an expectation of goodwill from another person. Consider two cases: in the first case, I trust you to transport an ornate tea kettle to the dining room. In the second case, I trust you to do the exact same thing, except now a number of my hyperactive baby cousins are zooming around the room at knee-height. Obviously I should trust you in the first case but not in the second -- after all, you have no clue what you're in for when these cousins arrive. Yet I rely on you in either case, and I do not doubt your goodwill toward me in either case. The action you perform in each case is the same. So, Baier's view by itself is insufficient for differentiating these two cases if we assume trust is a three-place relation, either holding or not for the triple $\langle\text{`me'}, \text{`you'}, \text{`carry the tea kettle'}\rangle$.

Karen Jones says that trust is an attitude of optimism about a trustee's goodwill and competence in the domain of interaction, such that the trustee is moved directly and favorably by the thought they are trusted. The idea of ``competence'' in the ``domain of interaction'' is telling.\autocite[4]{jones-affective-attitude} Do we take Jones to mean simply competence at the action? If so, then Jones' account does not seem equipped for the tea kettle example -- suppose I have no doubt in your ability to carry tea kettles under normal circumstances. The view can succeed if we take ``carrying a tea kettle while hyperactive children run around at knee-height'' as one kind of action, but this language is unnatural -- it does not feel characteristic of what an action is. If, instead, ``domain of interaction'' encompasses some facts about the situation in which the action takes place, then Jones is already hinting at context-sensitivity with this condition. On this reading, Jones' view seems to consider not whether $A$ trusts $B$ to do $X$, but rather whether $A$ trusts $B$ \textit{in domain of interaction $D$}, which looks similar to \textit{to do $X$ in context $C$}. Thus, adding context-sensitivity to a theory of trust, or finding out it was there all along, can help enrich the theory.

\subsubsection{``Anything Counts as Context'' Objection}
\textit{Objection}: In section \ref{ContextSensitivity} you stated that contextual factors can be facts about $A$ or various kinds of beliefs that $A$ has. These facts and beliefs could be just about anything! Since context can be so broadly construed, how can it possibly help us understand trust?

\textit{Response}: This objection may strike at some warranted discomfort with the word ``context,'' since it seems like context could just stand in to mean ``everything else that's important.'' Suppose we took this approach and left our contexts completely devoid of structure. Then, making trust a four-place relation could allow us to assess cases of trust with as much specificity as needed, but it would really add nothing informative. Fortunately, section \ref{ContextSensitivity} shows that this is not the case. Context $C$, like truster $A$ and trustee $B$, is constructed in a structured and restricted fashion, such that $C$ can only encode relevant facts about $A$, beliefs of $A$ about $B$, $X$, or past contexts in which $B$ attempted $X$ and the outcomes of those attempts. This set contains far less than the set of all facts about $A$ or all of $A$'s beliefs. Once again, the whole upshot of context-sensitivity is the difference-making feature of contextual factors. If some fact about $A$ will not make a difference to a particular case of trust, it should never appear in the context $C$.

To the point that contextual factors can be just about anything: well, yes. The condition that trust is \textbf{truster-subjective} (defined in section \ref{Directedness}) requires this flexibility. Since $A$ is the ultimate arbiter as to whether they trust $B$, then should $A$ say that the color of $B$'s shirt is relevant to their trust, we must include $B$'s shirt color in the context. If trust was a rational attitude, there could be more restriction on relevant contextual factors. However, people often trust and distrust irrationally, and any good philosophical theory of trust must account for these cases. Of course, if $B$'s shirt color was \textit{not} relevant to $A$'s trust, then it should not appear as a contextual factor, as argued for above.

\subsubsection{``Mere Reliance'' Objection}
\textit{Objection}: Your conditions for what it means to trust do not sufficiently distinguish trust from mere reliance, as theories like Annette Baier's, Karen Jones', or Richard Holton's manage to do. This makes the theory presented in \ref{DefiningContextualTrust} at least insufficient for capturing the nature of trust.

\textit{Response}: My defense from this objection hinges a bit on how successful C. Thi Nguyen's defense is of the normative force of his unquestioning attitude account. There are slight alterations, though, that I think make my theory more defensible.

Nguyen's defense of this objection considers what he calls the ``normative bite'' of our reactions to objects that we trust.\autocite[30]{Nguyen-trust} For Nguyen, one of the overarching purposes of trust is functional integration. In other words, just as in the Contextual Trust Account, we trust so to extend our agency in the world. In trusting we take objects, like cars and smartphones, to be extensions of ourselves and facilitators of our agency. When they fail, it is \textit{our} agency that is compromised. Sometimes such objects are deeply integrated in our agency at the time they fail, and the result can be quite alienating, closer to the loss of a limb than of some technological knickknack.

My reply to the objection begins with that exact reply from Nguyen above. Under the Contextual Trust Account, though, what it means to trust \textit{in the first place} is to already have identified $B$'s doing $X$ as in service to one's own ends. This differs slightly from the unquestioning attitude account, in which such functional integration is an important result of trusting that explains our willingness to do so. Instead, in the Contextual Trust Account, to trust \textit{at all} is to involve your trustee in your own project or goal. The ends-oriented focus on this account provides additional reason to suspect that contextual trust extends beyond reliance in important ways, even when the recipients of trust are non-human agents or objects.

\subsubsection{``Action Granularity'' Objection}\label{ActionGranularity}
\textit{Objection}: Under your view, context-sensitivity is required when specifying a truster, trustee, and action is not enough to disambiguate cases of trust. We could do away with context-sensitivity by instead considering fine-grained actions that include contextual factors. For example, ``driving in the city at night'' can be thought of as a fine-grained action and seems to specify contextual factors within the action. So, context-sensitivity does not seem as necessary as you claim.

\textit{Response}: This objection is tricky as its refutation could take us far afield. The granularity of actions\footnote{See \cite{anscombe_2000} and \cite{thompson_2012} for a primer.} and the logical form of action verbs\footnote{See \cite{kenny_1963} and \cite{davidson_2001}.} are entire subjects in the philosophies of action and language. It would be difficult to give a comprehensive treatment of the objection that satisfies both. Action verbs are particularly complicated in the philosophy of language. In English, we can add descriptive clauses to an action phrase indefinitely, making the action described steadily more granular. Consider ``Brutus killed,'' ``Brutus killed Caesar,'' ``Brutus killed Caesar in Pompey's theater,'' ``Brutus killed Caesar in Pompey's theater with a knife,'' and so on. As Anthony Kenny points out, we may be uncertain when to stop. And so ``if we cast our net widely enough, we can make `Brutus killed Caesar' into a sentence which describes, with a certain lack of specification, the whole history of the world.''\autocite[112]{kenny_1963}

So indeed, certain action terms may be suitably granular to have no meaningful context-sensitivity when it comes to trust. Yet we can observe that certain contextual factors, particularly the ones we have discussed, fit this schema more awkwardly than others. For example, take the second kind of contextual factors we covered: beliefs $A$ has about $B$ (in the case where $B$ does action $X$). It is awkward to say that ``$B$ does $X$ while $A$ believes $p$ about $B$'' describes a granular action of the more general form ``$B$ does $X$.'' At least, I believe speakers of English would not say this. $A$ believing $p$ really has nothing to do with $B$ \textit{doing the action $X$}. Instead, this rather passive fact affects how $A$ might trust $B$ to do $X$.

In fact, all of our contextual factors covered in section \ref{ContextSensitivity} are either facts about or beliefs of the truster $A$. Neither of these are relevant for the action, so it seems that context-sensitivity captures more than a more granular view of actions would allow. Failing this argument, it should be at least clear that context-sensitivity presents a more compact and manageable view for considering trust, so we may adopt it for pragmatic reasons.

\subsection{Case Study: Explainable Artificial Intelligence}\label{CaseStudyXAI}
\subsubsection{Trust and XAI}
Having finally established the Contextual Trust Account, we can turn to a case study in the field of Artificial Intelligence. Specifically, we examine a subfield of the discipline, Explainable Artificial Intelligence (henceforth XAI).

Recall from section \ref{AI} that XAI researchers consider trust to be a central goal of their work. Computer scientists Scott Lundberg and Su-In Lee claim that interpretability in model predictions ``engenders appropriate user trust.''\autocite[1]{LundbergL17} Been Kim calls transparency ``a major factor in establishing user trust in adaptive agents.''\autocite[102]{kim_2015} Dosilovic et al. indicate a growing necessity for trust in AI, as these technologies continue to encroach on our lives:
\begin{displayquote}
   ``Many things will be prescribed by such algorithms and that will affect human lives in ways maybe now unimagined so people will need to trust them in order to accept those prescriptions.''\autocite[210]{dosilovic}
\end{displayquote}

More specifically, Sundararajan et al. highlight the importance of trust in human-machine collaborative settings. Computer vision algorithms are applied in various medical settings to assist human professionals. For example, a deep network from Gulshan et al. uses computer vision to classify the severity of diabetic retinopathy, a complication of diabetes affecting the eyes, using retinal fundus images as input.\autocite{10.1001/jama.2016.17216} The algorithm was reviewed by a panel of ophthalmologists and achieves near-perfect classification accuracy. Sundararajan et al. comment of this model:
\begin{displayquote}
    ``Feature importance explanations are important for this network as retina specialists may use it to build trust in the network’s predictions, decide the grade for borderline cases, and obtain insights for further testing and screening.''\autocite[6]{DBLP:journals/corr/SundararajanTY17}
\end{displayquote}

Gulshan et al.'s model presents a compelling case for XAI, as their users are domain experts in the image classification task, yet (presumably) ignorant of modern machine learning. The model may classify images using a vastly different technique from professional ophthalmologists. Such collaboration can immensely improve patient outcomes, provided the professionals can trust the model's predictions enough to incorporate them into diagnoses. In this setting, trust, and \textit{not} model performance alone, is a key indicator of the model's real-world potential.

So XAI researchers have clearly identified the need to study trust. Unfortunately, these same researchers report a discordance in such study in their own literature. Dosilovic et al., in their survey on XAI in 2018, call trust ``hard to formalize and quantify,'' and hence ``usually criteria of interpretability and explainability are used as intermediate goals.''\autocite[210]{dosilovic}

Treating explanation as a proxy for trust is problematic for several reasons:
\begin{enumerate}
    \item The quality of an explanation says nothing about its indicating trust. A perfectly \textit{good} explanation can present a perfectly obvious reason to \textit{distrust} something. If ``I hit the hood with this wrench and it suddenly started up!'' is an accurate explanation for how I fixed your car, I am clearly a terrible car mechanic and should not be trusted with your vehicle. In other words, even perfect explainability is not sufficient for trust.
    \item Explanation is also not necessary for trust. Soldiers trust their platoon leaders without any explanation as to why some order was given. Children trust their caretakers and educators without explanation or understanding of their credentials. In these cases we trust by virtue of one's authority or demonstrated competence, and we need no explanation for either.
\end{enumerate}

Furthermore, traditional tripartite models of trust in philosophy are unlikely to help us. The point of XAI is not to determine whether a given individual trusts a given algorithm or not. Instead, XAI techniques actively attempt to influence user trust in cases where it may be absent, particularly where model complexity and opacity present barriers to trusting. This procedure -- changing trusting attitudes using auxiliary information -- clearly requires contextual factors. In the sections that follow we will show explicitly how XAI and contextual trust intertwine.

\subsubsection{The Problem Statement of XAI}
Modern artificial intelligence architectures, particularly deep neural network (DNN) architectures, generally perform better as their model complexity increases.\footnote{\cite[309]{carabantes2020}. Note that we deliberately avoid the technicalities of overfitting via overparameterization -- in fact virtually all state-of-the-art DNNs exhibit overfitting to various degrees without breaking this scaling law.} This basic scaling law has an intuitive explanation. Larger, more complex models have more learnable parameters, meaning they can store more nuanced internal representations of the problem they intend to solve, which increases accuracy. However, complexity often brings about \textbf{opacity}, that is, a general inability for humans to understand the model's inner workings and decision procedures.\autocite{burrell2016} Another term for opaque model architectures is ``black boxes.''\autocite[1]{ribiero-2016} According to researchers in XAI, opaque model architectures are undesirable because \textit{trust} in the models' predictions cannot be assessed,\autocite[1]{ribiero-2016,xai-wikipedia,dosilovic} leading to potential business, legal, and health hazards in cases of AI deployed in real-world contexts.\autocite{dosilovic} The phrase \textbf{trustworthy machine learning} is gaining popularity within XAI, and was even the name of a Stanford Computer Science course last year.\autocite{datta_mitchell}

At this point, many questions may be relevant to ask. \textit{Who} would we like to trust opaque AI systems -- AI practitioners themselves? End users, with little to no technical literacy? Stakeholders such as radiologists relying on AI-assisted decision-making in their work? And what are the varieties of trust that will suffice in each case? Also, is trust the correct criterion to use? Maybe we could consider, say, mathematical proof of the models' decision boundaries, or a guarantee of robustness against certain specific failure modes. Any of these dimensions could take us afield for the main task, though they represent promising future work. We will focus next on the proposed solutions from XAI.

\subsubsection{Explanation and Transparency}
Advocates of XAI propose techniques for algorithmic explanation and algorithmic transparency for the purpose of increasing trust in opaque AI systems.

\textbf{Explanations} consist of \textit{ex post}, human-interpretable descriptions illuminating how complex systems arrive at certain predictions from certain inputs.\autocite{DoshiVelez2017TheRO} The condition on human-interpretability is fluid, such that a description that counts as explanatory for one person may not count as explanatory for another. Consider, for example, AI scientists explaining their novel findings to one another at a conference. Their explanations are explanatory in context, but likely useless to anyone with a non-technical background.

\textbf{Transparency}, a related concept, measures the extent to which opaque systems can be understood via explanation. Kathleen Creel defines transparency, in this setting, as the inverse of opacity.\autocite[569]{creel2020} Transparency is a property of models themselves, whereas explanations attach themselves to certain procedures, such as model subroutines or the general functioning of a family of related models, such as the transformer architecture.\autocite{attention} There is therefore a certain respect in which one needs transparency to give good explanations. The philosophical literature on each is significant, though, and their relationship is not strictly as simple as necessity.

It is worth mentioning that in the AI literature, the supposed utility of explanation and transparency is not unanimously entertained. Zachary Lipton comments on the tenuous relationship between transparency and the very performance achievements that make AI systems so valuable to begin with:
\begin{displayquote}
    ``In some cases, transparency may be at odds with the broader objectives of AI (artificial intelligence). Some arguments against black-box algorithms appear to preclude any model that could match or surpass human abilities on complex tasks. As a concrete example, the short-term goal of building trust with doctors by developing transparent models might clash with the longer-term goal of improving health care.''\autocite[21]{lipton-mythos}
\end{displayquote}

XAI systems and techniques such as LIME,\autocite{ribiero-2016} QII,\autocite{QII} and SHAP\autocite{SHAP} take complex model behaviors and produce simplified explanations in human-interpretable formats. The medium can vary -- for example, LIME uses linear approximations of the base model's decision boundary near the prediction point, such that individual model features can be weighted by importance and reviewed by human interpreters. Other techniques use sections of images to report regions of interest in computer vision classifiers,\autocite{HendricksVisualExplanations} or explanations via counterfactual statements, like ``the model would have predicted \texttt{ACCEPT} had your reported annual income been \$10,000 higher.''\autocite{wachter}

The precise mathematical techniques and theory at work in these systems is regrettably out of scope here. There is a more general question in need of answer: how do we go about assessing model transparency in the first place, and how can transparency influence contextual trust?

\subsubsection{Modeling Transparency}
First, how can we model transparency in computational systems? Kathleen Creel has given a compelling account.\autocite{creel2020} According to Creel, there are three types of transparency in computational systems, mirroring David Marr's three levels of analysis.\autocite{marr}

First, \textit{algorithmic}, or \textit{functional} transparency, involves the high-level logical and computational rules of a system. In the case of a machine learning algorithm, this includes both the training procedure and the algorithm that is learned as a result of training. Algorithms are ``abstract mathematical objects,'' not their particular instantiations in code.\autocite[573]{creel2020} It is important to disambiguate functional transparency from another sense of the word ``functional,'' which concerns how constituent parts of a complex system fit together. Here, Creel means functional in the mathematical sense of function. Concretely, having functional transparency into an AI model might involve obtaining the \textit{pseudocode} of the inference algorithm, or a whitepaper giving an explanation of the training procedure, or a mathematical representation of the loss function, and so on.

Second, \textit{structural} transparency concerns how a particular algorithm is realized in code. Since algorithms are multiply realizable in a variety of different programming languages and styles, models with identical functional transparency may nonetheless admit different facts at the level of structural transparency. Structural transparency presents practical problems as the best modern computers can execute trillions of floating point operations per second (FLOPS), making real-time tracing of executable code practically impossible. According to Creel, what matters is the ``knowledge of relations between the subcomponents'' of a computational process, not the traversing of all possible paths.\autocite[578]{creel2020} This knowledge of subcomponent relationships sounds like ``functional'' knowledge in the previous sense we discussed -- the sense that is \textit{not} functional transparency. Concretely, having structural transparency into an AI model might involve obtaining the \textit{source code} of the model, plus documentation to explain the use of third-party functions imported into the code. Also, a diagram of different databases used, plus the linkages between them and the API specification for accessing them, contributes to structural transparency.

Third, and finally, \textit{run} transparency concerns how a program was run in a particular, individual instance. Transparency at this level concerns the hardware and input data upon which the model was run. Run transparency can capture artifacts in computational processes not visible at ``higher'' levels of abstraction, such as damaged hardware or distributional problems in input data. As a result, this level of transparency is the most specific to individual models and predictions by those models. Concretely, having run transparency into an AI model involves obtaining logs of individual training or inference runs, including what data was input to the model, what hardware ran the computation, and what results were output. Creel claims that physical access to the computational hardware can provide otherwise unattainable run transparency, such as corruption of sensitive detector equipment by cosmic rays.\autocite[580]{creel2020} However, it seems other forms of access, like obtaining a digital copy of the input data, can provide limited run transparency on their own.

Creel claims that these three levels of transparency are dissociable, and that statements at one level do not always entail statements at other levels. In certain settings, however, it is plausible that a model's source code, providing structural transparency, yields a generally acceptable description of the algorithm being computed, providing functional transparency. It seems less likely that facts at higher levels of abstraction entail facts at lower levels. Multiple realizability makes it hard to gain structural transparency from functional transparency, and run transparency considers factors just plainly missing at the higher levels, like input data and hardware.

\subsubsection{Transparency and Trust}
Creel's three levels of transparency in computational systems can help us understand the linkage between transparency and trust. In particular, Creel's taxonomy maps to a similar taxonomy of contextual factors for trust introduced in section \ref{ContextSensitivity}, including the truster's beliefs about the trustee, the action, and the trustee's past history of the action. With this taxonomy in hand, we can more clearly illustrate how transparency influences specific cases of contextual trust.

I argue that transparency fits into contextual trust as operations on the contextual belief set $C$. In other words, model transparency can influence trust via its ability to contribute \textit{difference-making beliefs} to an individual's decision to trust. This means that model explainability needs to be assessed \textit{independently} for each relevant truster, since the same explanation will affect different individuals differently. For example, a mathematics professor who receives functional information about an algorithm stands to gain a lot in terms of understanding the system's behavior. Yet the same functional information would be useless to someone without math education beyond trigonometry.

The first two layers, functional and structural transparency, correspond to contextual factors in $C$ of the second type explained in \ref{ContextSensitivity}. That is, functional and structural transparency admit explanations of model behavior that influence $A$'s (the truster's) beliefs about $B$ (the trustee, in this case the computational model). This formulation captures the condition that explanations cashed out of transparency at these levels affect different trusters differently. In our formulation of contextual trust, the second type of contextual factors in $C$ are \textit{beliefs that $A$ has about $B$} -- so, transparency is only worthwhile insofar as its ability to modify $A$'s belief set.

Next, the third layer, run transparency, corresponds to the third and fourth types of contextual factors in $C$. The third type includes beliefs $A$ has about the prediction task, $X$, which include facts about the input data used for the task. The fourth type includes observations, by $A$, of $B$'s ability to do $X$ in different contexts $C'$. In this case, the differing context $C'$ can contain facts about the particular hardware and input data used in previous model experiments. These members of $C$, as before, comprise the model's ``track record'' at a particular prediction task, which is exactly what run transparency is designed to explain.

With this taxonomical mapping, we can understand how transparency affects trust. Take a case where some person $A$ is deciding whether to trust model $B$ about some prediction task $X$. Transparency, in the form of explanations, operates on the context $C$ over which trust is evaluated. Suppose that $A$ is given transparency into the model $B$ via the following facts:
\begin{itemize}
    \item Model $B$'s runtime code is contained within this python file, \texttt{main.py}, which you can read. (structural transparency).
    \item For some individual $A'$ with a similar ZIP code, age, and gender to you, model $B$ ran the prediction task $X$ and arrived at some prediction $P$ (run transparency).
\end{itemize}
Once $A$ adopts these beliefs, the context $C$ transitions to a new context, $C'$, in which $A$'s new beliefs are contained. We can then question whether $A$ trusts $B$ to do $X$ in context $C'$.

Under our model for trust, explanation and transparency operate by changing contextual factors. We can rephrase the goal of XAI as (1) identifying contexts in which trust is absent and (2) augmenting said contexts with explanation and transparency such that trust is present. Say an individual $A$ (a person) does not trust $B$ (an algorithm) to do $X$ (a prediction task) in some context $C_{NT}$. Providing transparency, using XAI, \textit{changes the contextual conditions} from context $C_{NT}$ to some new $C_T$. If $A$ trusts $B$ to $X$ in context $C_T$, then our XAI technique was \textit{difference-making} to $A$'s trust, and hence successful.

\subsubsection{Concrete Example}
To present a concrete example of this connection at work, we return to Sundararajan et al.'s analysis of Gulshan et al.'s computer vision model. This model is a binary classification model using retinal fundus images (images of the back surface of the eye). This means the model takes, as input, an $n\times n$ matrix of pixels $M$, where each $p_{i, j}\in M$ is a triple $\langle r, g, b\rangle$ of integer values between 0 and 255. The model outputs a probability $\hat{y}\in[0, 1]$ representing the model's confidence that the image in question depicts diabetic retinopathy. We may suppose that values of $\hat{y}$ above 0.5 indicate prediction of the condition, though model users may have access to the actual predicted probability as a measure of the model's ``confidence.'' So, we can represent the model as a function $F: ((\mathbb{N}\cap [0, 255])^3)^{n\times n}\to [0, 1]$ mapping pixel matrices to predictions.

The model's architecture is an alteration of the Inception-v3 architecture for computer vision,\autocite{DBLP:journals/corr/SzegedyVISW15} consisting of a densely connected network of matrix convolution operators and normalization layers. Such an architecture basically runs a long and complicated series of matrix transformations on the original input $M$, producing intermediate representations of the image of varying sizes. These intermediate representations are pooled together to ultimately compute the desired prediction $\hat{y}$. Such a model may contain hundreds of thousands of trainable parameters, and execute billions of floating point operations in the process of training those parameters. This is far beyond a human's capacity to trace precisely, and so even the model's architects can have minimal understanding of how the model derives predictions. The models are, in other words, black boxes.

The prediction task at hand is the diagnosis of diabetic retinopathy given retinal fundus images from a patient with diabetes. Trained ophthalmologists typically identify diabetic retinopathy from the particular pattern of lesions left on the retina. In a panel survey, Gulshan et al. found that such ophthalmologists agreed with their model's predictions over 94\% of the time.\autocite{10.1001/jama.2016.17216} So, it is plausible that the model is learning to identify lesions in the images and predict depending on their features. Yet, without insight into the model's internal representations, it is impossible to know this for certain. If the model's decision procedure is completely unknown, it seems likely that ophthalmologists will not trust the model's predictions in certain edge cases where their intuition clashes, despite the model's high accuracy on average.

To provide transparency into Gulshan et al.'s model, Sundararajan et al. propose an XAI method called ``feature attribution with integrated gradients.''\autocite{DBLP:journals/corr/SundararajanTY17} This technique begins by selecting a \textbf{baseline input}. In the case of computer vision, this is typically an $n\times n$ matrix $B$ of all black pixels or noise, such that the model's prediction $F(B)\approx 0.5$. In other words, the model should be neutral as to whether the baseline input has a positive or negative prediction. Next, the technique computes a \textbf{salience score} for each pixel in the actual input image $M$, defined basically as a quantitative measure of that pixel's contribution to the predicted $\hat{y}$, normalized by the pixel's deviation from the matching pixel in the baseline $B$. Finally, the technique identifies neighborhoods of pixels with high salience scores and annotates these as either positive or negative attributes of the image, depending on their effect on the prediction score. In Gulshan et al.'s model, integrated gradients tend to pick up retinal lesions as both positive and negative attributes, validating the hypothesis that the model is analyzing the features of these lesions to form its prediction.

How can Sundararajan et al.'s integrated gradients approach inform trust in Gulshan et al.'s model? Let's consider an example case where one is deciding to trust a particular model prediction, $\hat{y}^*$, given input image $M^*$. Suppose $A$ is our agent who wishes to trust the model, $B$, and $X$ is the diabetic retinopathy prediction task. Suppose the following facts are available:
\begin{enumerate}
    \item On previous input images $M_1,\dots,M_k$, the model gave predictions $\hat{y}_1,\dots,\hat{y}_k$ (run transparency). Suppose we can visually examine each input image $M_i$.
    \item For our particular image $M^*$, the integrated gradients technique identified clusters of pixels $p_a, p_b, \dots$ as \textit{positive attributes} for the prediction, and clusters $n_1, n_2, \dots$ as negative attributes for the prediction. In other words, clusters $p_a, p_b, \dots$ contributed to the prediction $\hat{y}^*$ being closer to 1, and $n_1, n_2, \dots$ contributed to the prediction being closer to 0. These facts encode both functional transparency, since they specify part of the model's decision procedure, and run transparency, since they concern features of the input data. We can also suppose we have the same kind of cluster information for previous input images $M_1,\dots,M_k$. For our image $M^*$, the positive attributes $p_a, p_b, \dots$ look visually like lesions or other artifacts on the image of the retina.
    \item On previous input images $M_1,\dots,M_k$, a panel of professional ophthalmologists agreed with the model's predictions $\hat{y}_1,\dots,\hat{y}_k$ over 90\% of the time.
\end{enumerate}
These facts correspond to the following contextual factors, which are beliefs of the truster $A$, and together make up our current context $C$:
\begin{enumerate}
    \item In past contexts $C',C'',\dots$, the model $B$ did $X$ and arrived at the outcomes $\hat{y}_1,\dots,\hat{y}_k$. These past contexts might resemble the current context, $C$, to varying degrees.
    \item Model $B$ appears to use the presence of particular lesions or other artifacts in the retinal image for making its prediction, $\hat{y}^*$.
    \item In past contexts $C', C'',\dots$, the model $B$ appears to be generally in agreement with the expert human opinions on $X$, the diabetic retinopathy prediction task.
\end{enumerate}
Notice the important condition that these latter three points, not the former ones, count as contextual factors. These latter three points are all \textit{beliefs that $A$ has}, whereas the former three points are objective facts. Keeping contextual factors as beliefs of $A$ is important, since different trusters $A$ will respond differently to acquiring these beliefs. For example, a layperson (like myself) is likely to learn very little from point 2, concerning the particular lesions that were relevant for prediction. I know nothing about retinal lesions. I probably could not tell them apart from motes of dust on the lens, nor could I tell which lesions were indicative of the condition. In this case, I am likely to lean heavier on point 3, regarding the past consensus by experts, when determining my trust. However, a different truster $A$ -- an ophthalmologist, perhaps -- may decide to trust the algorithm or not depending on condition 2. If the model identifies the ``right kind'' of image features in making its prediction, according to a professional's domain knowledge, then the professional can more readily trust the prediction. Since point 2 is the only one capturing the effects of our XAI intervention, integrated gradients, this example shows how successful explanation is not always a guarantee for trust. The difference-makers, again, are the difference-making beliefs that $A$ has after the XAI information becomes available.

\subsubsection{Implications for XAI}
As our concrete example above illustrates, the takeaways for XAI from this exercise are relatively straightforward.

First, the Contextual Trust Account incorporates transparency and explanations as operations on contextual belief sets ($C$ in our formalization). Gaining transparency into a model changes a potential truster's beliefs about that model. As a result, the truster can consider trust in a \textit{new} context -- hopefully a context where trust is more appropriate than it was before. If XAI researchers adopt the Contextual Trust Account, then their goal is twofold: (1) identify contexts in which trust is absent, and (2) augment the truster's beliefs with explanation and transparency such that trust is present.

Second, there is a taxonomic mapping from types of transparency in computational systems to types of factors relevant for contextual trust. Functional and structural transparency influence truster $A$'s beliefs about the model $B$, and run transparency influences $A$'s beliefs about the prediction task $X$ and $B$'s track record at performing $X$ under different circumstances. Different XAI techniques target different levels of transparency, which means they likewise seek to influence trust in correspondingly different ways.

Third, the objective facts presented by XAI do not influence trust directly. These facts affect trust only in their ability to inspire difference-making beliefs in the relevant trusters. For this reason, XAI techniques may be useful only for particular types of end users, as groups like laypeople and medical professionals are likely to form very different beliefs from the same evidence. XAI researchers seem to grasp this point implicitly, as many include user studies in their results that demonstrate an alignment between their techniques and human intuitions.\autocite{10.1001/jama.2016.17216,LundbergL17,DBLP:journals/corr/SundararajanTY17} Some social scientists, like Papenmeier et al., have taken to studying the effect of explanations on self-reported trust empirically.\autocite{papenmeier} I have hoped to demonstrate how a contextual view of trust can help accommodate these studies in philosophical analysis, using the concept of difference-making beliefs in particular. The Contextual Trust Account is one candidate for the formal criteria of trust sought after by Dosilovic et al. and other XAI researchers.

\section{Conclusion}
This thesis advanced three arguments. The first argument intended to convince you that trust in philosophy ought to be understood contextually. We illustrated commonsense examples of trust in which the presence of contextual factors, beyond describing the trusters, trustees, and actions, can be difference-making factors. We then surveyed the existing philosophical literature on trust. In Section \ref{Background} we surveyed the past literature, identifying the dominant paradigm of interpersonal trust and exploring the most popular theories, like Baier's\autocite{baier_1986}. We then turned to impersonal theories and the growing body of work there, most notably Walker's\autocite{walker_2006}, Buechner and Tavani's\autocite{buechner-tavani}, Taddeo's,\autocite{taddeo_2010} and Nguyen's\autocite{Nguyen-trust}.

In Section \ref{ContextualTrust} we dug into the formulation of contextual trust with the Contextual Trust Account. Our second argument intended to convince readers that we trust in order to receive unexamined assistance in the projects we undertake. This agencial integration and intimacy can explain the ``normative bite'' that accompanies the loss of trust, distinguishing trust from mere reliance, while still allowing non-human entities like computational systems and personal devices to play the role of trustees.

Finally, we leveraged the Contextual Trust Account to explain a relevant, contemporary case of trust in artificial intelligence. Explainable Artificial Intelligence defines the task of building ``trustworthy'' AI systems, without a good understanding of how the concept of trust applies to such systems. Missing from XAI is a robust theory of trust that is calculable, philosophically acceptable, and respecting of the minimal conditions under which a trusting instance can be decided. In section \ref{CaseStudyXAI}, we explored how the Contextual Trust account can accommodate transparency as particular kinds of contextual factors, and argued that the goal of XAI should be to present difference-making factors of this kind to relevant users. Contextual trust can thus be of relevance to computer scientists, XAI researchers, philosophers, and anyone generally interested in studying the behavior of complex, collaborative systems.

\newpage
\printbibliography
\end{document}